\documentclass[10pt,twocolumn,letterpaper]{article}
\newcommand{\confName}{}
\usepackage{threeparttable} % for table notes
\usepackage{booktabs}       
\usepackage[pagenumbers]{cvpr} % To force page numbers, e.g., for an arXiv version
\usepackage{longtable}
\usepackage{tabularx}
\usepackage[most]{tcolorbox}
\usepackage{xcolor}
\usepackage[margin=1in]{geometry}
\usepackage[T1]{fontenc}
\usepackage[utf8]{inputenc}
\usepackage{booktabs}
\usepackage{multirow}
\usepackage{siunitx}
\usepackage[table]{xcolor}
\usepackage{caption}
\usepackage{graphicx}
\usepackage{geometry}
\usepackage{fontawesome}
\newtcolorbox{RoundedTitleBox}[2][]{%
  enhanced,
  breakable,                % 支持分页（若内容很长）
  colback=white,            % 正文背景
  colframe=black,           % 边框颜色
  colbacktitle=black,       % 标题背景色（黑）
  coltitle=white,           % 标题文字色（白）
  fonttitle=\bfseries,      % 标题字体样式
  colupper=black,           % 正文字体颜色（黑）
  title={#2},               % 标题文本（作为第二个参数传入）
  arc=4mm,                  % 圆角半径，修改此值调整圆角大小
  boxrule=0.8pt,            % 边框粗细
  left=6pt,right=6pt,top=6pt,bottom=6pt, % 内边距
  halign title=center,      % 标题居中（可改为 left/right）
  #1                        % 允许通过可选参数覆盖上述设置
}
\definecolor{cvprblue}{rgb}{0.21,0.49,0.74}
\usepackage[pagebackref,breaklinks,colorlinks,allcolors=cvprblue]{hyperref}

\def\paperID{39936} % * Enter the Paper ID here
\def\confName{CVPR}
\def\confYear{2026}

\title{Step-CoT: Stepwise Visual Chain-of-Thought for Medical Visual Question Answering}

\author{Lin Fan$^1$, Yafei Ou$^{2,3}$
\thanks{Corresponding author: Yafei Ou (\tt\small yafei.ou@riken.jp)}, Zhipeng Deng$^4$, Pengyu Dai$^{2,5}$, Hou Chongxian$^6$, Jiale Yan$^7$, \\Yaqian Li$^7$, Kaiwen Long$^7$, Xun Gong$^1$, Masayuki Ikebe$^3$, Yefeng Zheng$^4$\\
$^1$ Southwest Jiaotong University\\
$^2$ RIKEN\\
$^3$ Hokkaido University\\
$^4$ Westlake University\\
$^5$ The University of Tokyo\\
$^6$ Shenzhen People’s Hospital\\
$^7$ Li Auto Inc.
}
\lstdefinelanguage{json}{
    basicstyle=\ttfamily\small,
    breaklines=true,
    breakatwhitespace=true,
    showstringspaces=false,
}
\begin{document}
\maketitle

\begin{abstract}
Chain-of-thought (CoT) reasoning has advanced medical visual question answering (VQA), yet most existing CoT rationales are free-form and fail to capture the structured reasoning process clinicians actually follow. This work asks: Can traceable, multi-step reasoning supervision improve reasoning accuracy and the interpretability of Medical VQA?
To this end, we introduce Step-CoT, a large-scale medical reasoning dataset with expert-curated, structured multi-step CoT aligned to clinical diagnostic workflows, implicitly grounding the model’s reasoning in radiographic evidence. Step-CoT comprises more than 10K real clinical cases and 70K VQA pairs organized around diagnostic workflows, providing supervised intermediate steps that guide models to follow valid reasoning trajectories.
To effectively learn from Step-CoT, we further introduce a teacher-student framework with a dynamic graph-structured focusing mechanism that prioritizes diagnostically informative steps while filtering out less relevant contexts. Our experiments show that using Step-CoT can improve reasoning accuracy and interpretability.
  \\
  \small \textbf{\mbox{\faGithub\hspace{.25em} Benchmark:}} \href{https://github.com/hahaha111111/Step-CoT}{github.com/hahaha111111/Step-CoT}\\
  \raisebox{-0.3\height}{\includegraphics[width=0.35cm]{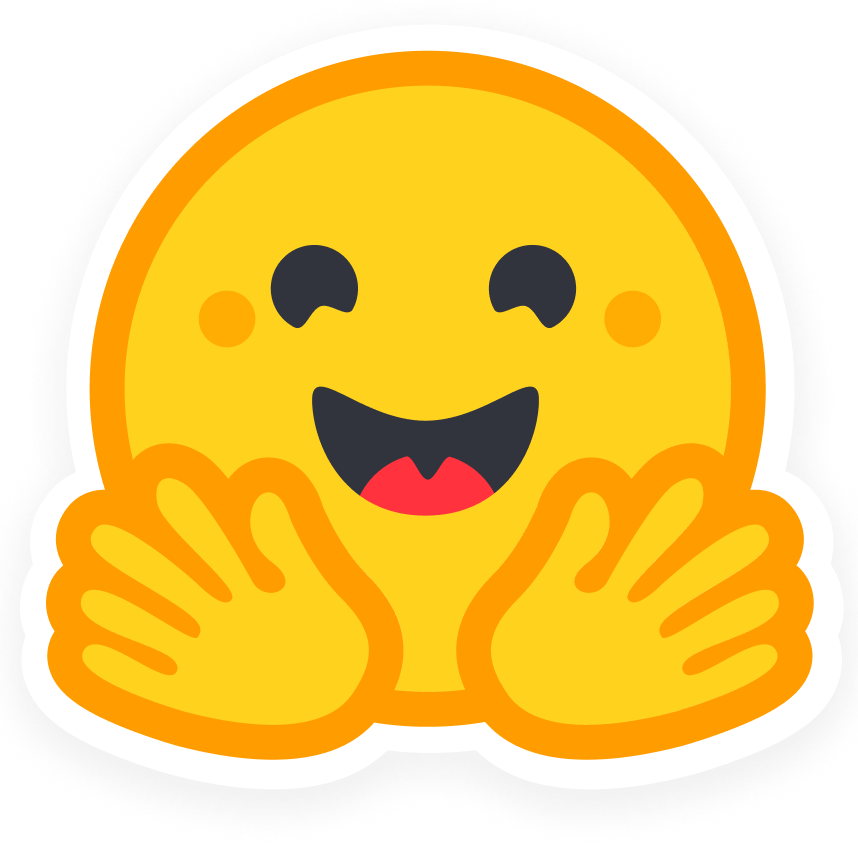}} \small \textbf{\mbox{Dataset Card:}} \href{https://huggingface.co/datasets/fl-15o/Step-CoT}{huggingface.co/datasets/fl-15o/Step-CoT}
\end{abstract}

\begin{table*}[!t]
  \centering
  \caption{Comparison of recent medical reasoning datasets with Step-CoT.}
  \resizebox{\textwidth}{!}{%
  \begin{tabular}{ccccc}
    \toprule
    Dataset & Instances & CoT Construction & Expert Involvement & Clinical Workflow \\
    \midrule
    PMC-VQA~\cite{zhang2023pmc}  & 227K &\textcolor{red}{$\times$}No CoT &\textcolor{green}{$\checkmark$}Yes & \textcolor{red}{$\times$}No \\
    Med-GRIT-270k~\cite{huang2024refer} &270k QA pairs &\textcolor{green}{$\checkmark$}GPT-generated&\textcolor{red}{$\times$}No experts &\textcolor{red}{$\times$}No\\
    MedCoT~\cite{liu2024medCoT}  & 50k Extends Med-VQA & \textcolor{green}{$\checkmark$}Expert validation & \textcolor{green}{$\checkmark$}Yes & \textcolor{orange}{$\checkmark$}Partial \\
    ReasonMed~\cite{cai2025reasoning} &370k reasoning samples &\textcolor{green}{$\checkmark$}Agent-generated &\textcolor{green}{$\checkmark$}Multi-agent validation &\textcolor{red}{$\times$}No\\
    HVCR~\cite{ding2025building} &31k QA pairs&Expert validation &\textcolor{green}{$\checkmark$}Yes &\textcolor{red}{$\times$}No\\
    V2T-CoT~\cite{wang2025v2t} &39k examples &\textcolor{green}{$\checkmark$}GPT-generated CoTs &\textcolor{red}{$\times$}No experts&\textcolor{red}{$\times$}No\\
    MedThink~\cite{gai2025medthink} &50k Extends Med-VQA &\textcolor{green}{$\checkmark$}Decision-making rationales &\textcolor{green}{$\checkmark$}Semi-auto + human validation &\textcolor{red}{$\times$}No \\
    % \midrule
    Step-CoT (Ours) & 70K QA pairs with reasoning& \textcolor{green}{$\checkmark$}Structured multi-step &\textcolor{green}{$\checkmark$}Semi-auto + human validation & \textcolor{green}{$\checkmark$}Yes \\ 
    \bottomrule
  \end{tabular}}
  \label{tab:CoT_datasets_comparison}
\end{table*}

\section{Introduction}
\label{sec1}
Medical Visual Question Answering (Med-VQA) has emerged as a critical topic in healthcare AI, leveraging multi-modal deep learning to answer clinical questions related to images posed in natural language~\cite{liu2025gemex,dai2024sasamim,fan2024tri,fan2025cycle,dai2025goca,fan2026evolving}. Leveraging the ability to generate coherent responses while incorporating extensive medical domain knowledge, Med-VQA has demonstrated practical utility across diverse tasks such as computer-aided diagnosis and fine-grained tumor attribute recognition~\cite{liu2025gemex,zhan2020medical,gong2022vqamix}. Recent advances in Med-VQA models have evolved from focusing on performance enhancement through scaling model architectures and expanding pre-training datasets~\cite{yang2025ram}, such as ExGra-Med~\cite{nguyenenriched}, LLaVA-Med~\cite{li2023llava}, MedGemma~\cite{sellergren2025medgemma}, and LLaVA-Tri~\cite{xie2024medtrinity}, to improving interpretability by decomposing complex diagnostic tasks into step-by-step reasoning processes through CoT mechanisms~\cite{zhao2025cot,xu2025llava}, including ReasonMed~\cite{cai2025reasoning}, HVCR~\cite{ding2025building} and MedCoT~\cite{liu2024medCoT}. By explicitly revealing their intermediate reasoning steps, CoT methods enhance both predictive accuracy and interpretability, rendering them particularly valuable for high-stakes domains such as healthcare.

Recent studies have explored automatic generation of CoT reasoning data for Large vision language models (LVLMs) to improve reasoning accuracy and enhance interpretability. Approaches such as MedCoT~\cite{liu2024medCoT}, MedThink~\cite{gai2025medthink}, ReasonMed~\cite{sun2025reasonmed}, and the HVCR~\cite{ding2025building} provide textual reasoning traces that enable models to produce rationales alongside predictions. To further connect reasoning with visual evidence, datasets like V2T-CoT~\cite{wang2025v2t}, Med-GRIT-270k~\cite{huang2024refer}, and MedTrinity-25M~\cite{xie2024medtrinity} pair CoT statements with image annotations (as shown in Table~\ref{tab:CoT_datasets_comparison}).
Although these approaches improve data availability, their effectiveness in modeling clinical reasoning remains limited by two major issues:
(i) These datasets lack a structured, stepwise diagnostic protocol. They either provide free-form rationales or automatically generated reasoning chains that fail to align with real clinical workflows and omit intermediate diagnostic states reflecting radiologists’ sequential decision-making. 
(ii) Most CoT datasets rely heavily on GPT-4.1-based synthetic rationales derived from existing image-text pairs, raising serious concerns about factual inconsistency.

Beyond dataset construction, training paradigms for CoT in LVLMs further enlarge this gap. Most CoT training paradigms for LVLMs, which rely on Supervised Fine-Tuning (SFT) or Reinforcement Learning (RL), are inherently non-interactive and perceptually static~\cite{ziegler2019fine,chu2025sft}, primarily using a static image plus question input and reporting final-answer outputs without explicit action-level reasoning chains~\cite{li2023llava}, limiting the model to the initial input image and preventing it from actively gathering new information or refining its perception. For example, BLIP‑2~\cite{li2023blip} has achieved strong open‐domain VQA performance but lacks mechanisms for interactive perceptual refinement or multi‐step visual reasoning, while emerging specialized medical LVLMs such as LLaVA‑Med~\cite{li2023llava} demonstrate domain adaptation but still without action-level reasoning chains~\cite{li2023llava}. Similarly, the recent MedVLM‑R1~\cite{pan2025medvlm} framework uses RL to incentivize reasoning paths in medical image analysis. Yet, the perceptual input remains static, and the model cannot execute intermediate actions that change what it perceives.

These limitations underscore the need for a structured reasoning annotation framework that not only integrates visual reasoning and aligns multi-step inference with multi-modal clinical evidence, but also enables traceable, stepwise reasoning interactions with visual data, allowing each reasoning step to dynamically update and refine diagnostic understanding in a clinically consistent manner. This leads to the central question of this work:\\
\emph{Can traceable, multi-step reasoning supervision improve reasoning accuracy and the interpretability of Medical VQA?}

To solve this problem, we introduce Step-CoT, a novel medical dataset for multi-step vision-language reasoning. Step-CoT organizes clinical diagnostic workflows with clinician-curated intermediate reasoning steps, providing the necessary supervision for models to learn dynamic, multi-step reasoning through actionable sequences that can alter perceptual input. By structuring reasoning as a series of clinical actions, our dataset facilitates a transition from static reasoning to a multi-step problem-solving paradigm. Along with the dataset, we release benchmark evaluations, pretrained baselines, and a teacher-student framework to support supervised intermediate reasoning and context-conditioned inference.
To summarize, this paper makes the following contributions:
\begin{itemize}
    \item We present Step-CoT, a dedicated medical visual CoT dataset comprising more than 10K real clinical cases and 70K question-answer pairs. Each instance contains clinically grounded reasoning chains. Our experiments show that Step-CoT progressively aligns multi-step reasoning with visual evidence, guides models to follow valid and traceable diagnostic trajectories, enhancing the accuracy, interpretability, and clinical relevance of Med-VQA.
    \item To effectively learn from Step-CoT, we propose an innovative teacher-student CoT reasoning framework that distills complex clinical reasoning knowledge from the dataset into a lightweight student model, thereby improving generalization and adaptability to diverse diagnostic tasks.
    \item We introduce a visual CoT benchmark for Med-VQA that evaluates performance in scenarios requiring clinically faithful reasoning chains and evidence-based justifications for supporting diagnosis.
\end{itemize}

\begin{figure*}
    \centering
    \includegraphics[width=\linewidth]{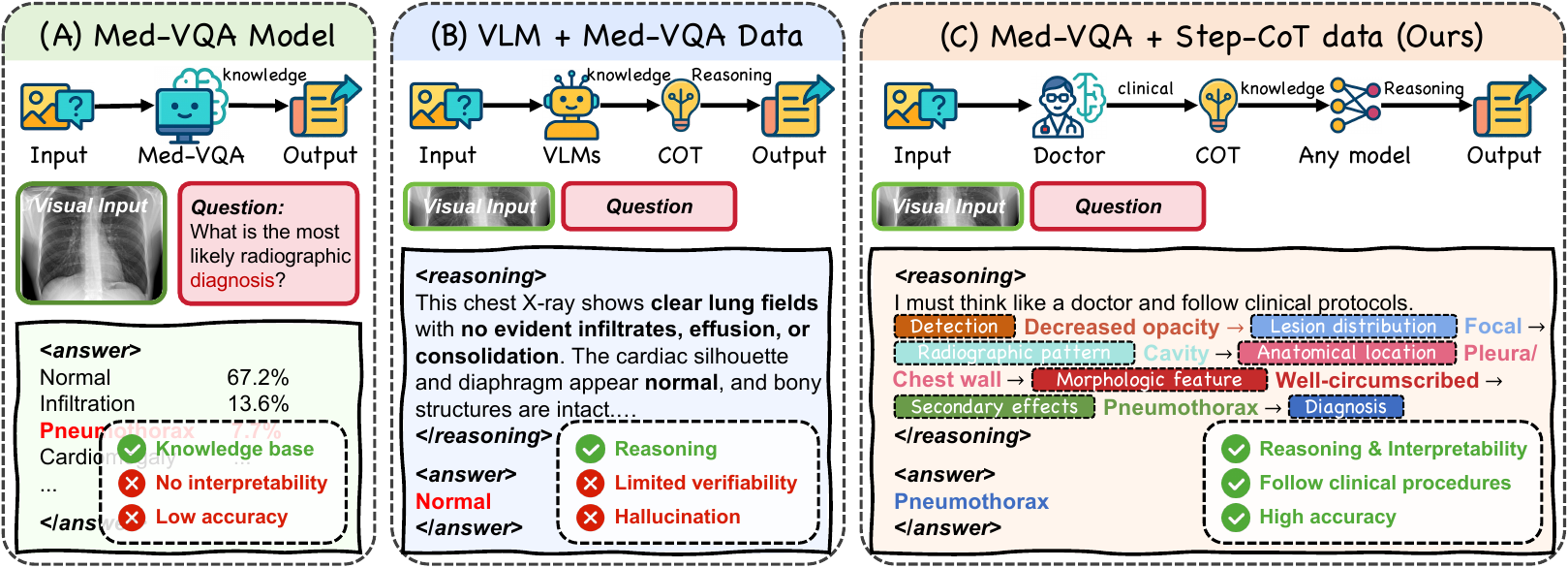}
    \caption{Overview of the Step-CoT dataset. (A) Conventional Med-VQA approaches, where models take an image and a question as input, perform multi-modal feature fusion and output a diagnostic answer. Although leveraging multi-modal knowledge, this paradigm lacks interpretability and often yields limited diagnostic accuracy. (B) Enhances interpretability by integrating large language models with CoT reasoning to generate intermediate explanations; however, such reasoning is often unreliable. (C) Our proposed Step-CoT dataset and training framework, which introduces explicit intermediate supervision. By guiding the model to learn structured clinical reasoning steps, Step-CoT not only improves interpretability through trustworthy intermediate reasoning but also enhances diagnostic accuracy.}
    \label{fig:1}
\end{figure*}

\section{Step-CoT Dataset}
\label{sec:dataset}

We constructed the medical visual CoT dataset Step-CoT (see Fig.~\ref{fig:1}). Step-CoT formalizes the clinical diagnostic trajectory as a seven-step, sequential reasoning process and applies full supervision across the entire diagnostic pipeline, including ground-truth answers and intermediate reasoning annotations for each step. Each sample comprises one medical image, seven clinically relevant reasoning questions, and their corresponding answers with intermediate supervisory signals. Every answer is accompanied by an explanatory reasoning chain that articulates the thought process behind it.

\subsection{Data Collection}
This work comprises original Chest X-ray (CXR) images and diagnostic text drawn from three public sources (totaling 10,068 CXR studies):
(i) IU X-Ray~\cite{demner2016preparing}, we use the subset of 3749 CXR studies.
(ii) PadChest-GR~\cite{castro2024padchest}, we use the subset of 3230 CXR studies.
(iii) Med-Image-Reports~\footnote{\url{https://huggingface.co/datasets/zirui3/med-image-reports}}, we use the subset of 3089 CXR studies. 
Appendix Sec. B provides more detailed information regarding the dataset, including the sample selection criteria and dataset structure, among other details.

\subsection{Data Annotation Process}
This study employed a structured analytical framework to evaluate chest X-ray radiology reports using DeepSeek-R1, an LLM that has demonstrated robust performance in clinical reasoning tasks~\cite{sandmann2025benchmark,tordjman2025comparative}. For annotation, we first collected the original clinician-authored, free-text radiology reports provided with the datasets. These unstructured reports were submitted to DeepSeek-R1 (prompts are detailed in Appendix Sec. A) to extract structured image findings and attributes. The model outputs were then mapped onto the predefined multi-step reasoning schema, producing per-step question-answer entries and explicit reasoning chains. The dataset consisted of systematically processed radiology reports paired with chest X-ray images, ensuring comprehensive case representation. Each case included three structured components: a sequential set of clinical questions, corresponding answer logic, and explicit reasoning chains. This design provided the methodological foundation for evaluating both diagnostic accuracy and interpretability in medical visual question answering. Table~\ref{tab:case_components} summarizes the structured components of the analysis framework.

\begin{table}[!t]
\centering
\caption{Components of individual case analysis}
\label{tab:case_components}
\resizebox{\linewidth}{!}{%
\begin{threeparttable}
\begin{tabular}{p{1.8cm}p{7cm}}
\toprule
{Component} & {Description and Implementation} \\
\midrule
Clinical Questions & Sequential questions derived from radiological diagnostic pathways, progressing from abnormality detection to final diagnosis. \\
\midrule
Answer Logic & \begin{itemize}[leftmargin=*, nosep, itemsep=0pt, before=\vspace{-2.5mm}]
\item Direct answers from explicit report content.
\item ``N/A'' when the question is clinically irrelevant.
\item ``No Answer'' when information is missing, with inference attempted from available evidence.
\end{itemize} \\
\midrule
Reasoning Chains & Each reasoning step required explicit justifications derived directly from report content or clinically valid inferences. This requirement ensures that model outputs remain grounded in available clinical evidence while maintaining the structured analytical rigor characteristic of expert radiological interpretation. \\
\bottomrule
\end{tabular}
\end{threeparttable}%
}
\end{table}

The case analysis protocol was constructed to emulate established radiological diagnostic workflows~\cite{de2012chest,scott2004lung,bansal2019interpreting}. For each patient case, clinical questions progressed in a stepwise manner, beginning with abnormality detection, moving through pattern characterization and spatial assessment, and concluding with diagnostic synthesis. This seven-step cascade ensured that each analytical stage logically built upon prior conclusions, thereby maintaining clinical coherence and mirroring the reasoning structure employed by expert radiologists. The prompt engineering implemented a seven-step analytical cascade that maintains contextual continuity across diagnostic stages:

\begin{itemize}
   \item \textbf{Abnormal Radiodensity Detection (Detection step)} Determine the presence or absence of abnormal radiodensity in the lungs or surrounding thoracic structures (e.g., increased, decreased, or mixed opacity).

    \item \textbf{Appearance Survey (Lesion distribution step and Radiographic pattern step)} Characterizes any detected abnormality by assessing its spatial distribution (e.g., focal, diffuse) and its predominant basic radiographic pattern (e.g., reticular, consolidation).

    \item \textbf{Feature Analysis (Anatomical location step, Morphologic feature step, and Secondary effects / associated signs step)} Refines the description by specifying the precise anatomical location of the lesion (e.g., right upper lobe), its margins and internal morphology (e.g., well-circumscribed), and any secondary effects on surrounding structures or lung volumes (e.g., mediastinal shift).

    \item \textbf{Diagnostic Synthesis ({Diagnosis step})} Integrates all previous findings to formulate a comprehensive radiographic diagnosis or impression (e.g., atelectasis).
\end{itemize}

This cascading reasoning structure ensures that each analytical step logically builds upon previous conclusions, creating a coherent diagnostic narrative that reflects expert clinical thought processes. The prompt was specifically instructed to reference prior step conclusions in subsequent reasoning, thereby maintaining diagnostic continuity and reducing contextual fragmentation. All diagnostic procedures and data labels were reviewed and verified by a board-certified physician, who obtained his Medical Practitioner Certificate in 2018.

\begin{figure*}[!t]
    \centering
    \includegraphics[width=0.95\linewidth]{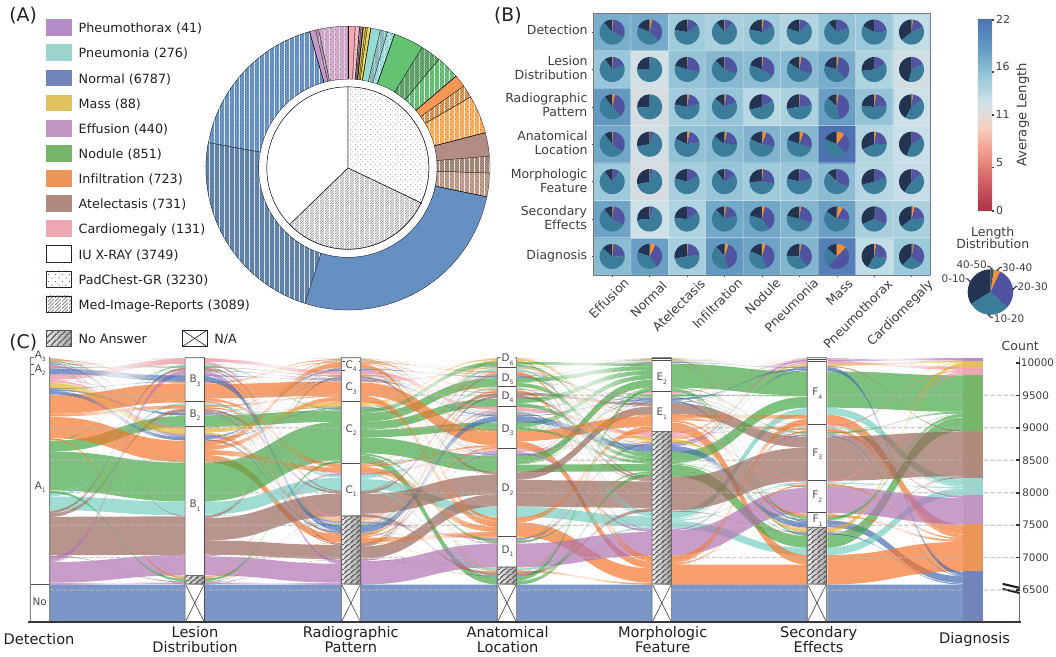}
    \caption{Distribution and statistics for the data sources, disease prevalence, answer distributions, and reasoning lengths in the Step-CoT dataset. (A) The inner ring illustrates the proportional distribution across different datasets, while the outer ring represents the distribution of various disease categories within the datasets. (B) This confusion matrix, organized by disease categories and reasoning steps, visualizes the average reasoning chain length. Each cell contains a pie chart representing the statistical distribution of samples across different chain lengths, while the marginal histograms on the axes display the sample count distributions by chain length for individual steps (x-axis) and disease categories (y-axis). (C) This diagram presents the outcome transition statistics between consecutive reasoning steps, mapping the flow of diagnostic conclusions throughout the clinical reasoning pathway. The name of each annotation (e.g., A1) can be referred to in the dataset description section of Appendix Sec. B.}
    \label{fig:analysis}
\end{figure*}

\subsection{Statistics of Step-CoT}
We performed a statistical analysis of the Step-CoT dataset, examining data sources, disease prevalence, answer distributions, reasoning lengths, and other key attributes. Detailed statistics are summarized in Fig.~\ref{fig:analysis}. From the diagnostic-label perspective (Fig.~\ref{fig:analysis}~(A)), IU-CXR is enriched with normal, effusion, and cardiomegaly cases; BIMCV is dominated by Normal and Nodule cases; and Med-Image-Reports exhibits a comparatively broader spread, including less frequent conditions such as mass and pneumothorax. These inter-dataset differences, together with step-level applicability signals, provide a valuable substrate for assessing model robustness and generalizability under varying label priors and reporting styles. Data drawn from the three sources show a consistent pattern: instances labeled Normal constitute more than 50\% of each cohort, while other disease categories are relatively balanced. This class imbalance has been widely documented in clinical cohorts and aligns with the clinical findings reported by Alshanketi~\cite{alshanketi2025pneumonia}. Preserving this natural imbalance, rather than performing artificial class balancing, can help avoid a common pitfall in AI studies and ensure that training and evaluation results retain clinical relevance.

In terms of textual complexity (Fig.~\ref{fig:analysis}~(B)), the mean word count per reasoning step varies across datasets (IU-CXR: 15-19 words; Hugging: 15-25 words; BIMCV: 9-11 words), indicating heterogeneous linguistic complexity and reasoning depth. The modest increase in sentence length toward later steps suggests that more elaborate, conclusive statements tend to be produced after successive reasoning stages. 

Analysis of the multi-step answer distribution across the dataset (Fig.~\ref{fig:analysis}~(C)) indicates that many samples follow relatively consistent reasoning trajectories, in which a limited number of principal diagnostic flows account for a substantial portion of reasoning transitions. This pattern suggests that the seven-step question-answer sequences tend to form stable, clinically coherent pathways that broadly align with structured diagnostic reasoning processes observed in clinical practice~\cite{detterbeck2013screening}. Notably, the dominance of reasoning flows implies that the final diagnostic conclusions are strongly conditioned by earlier reasoning steps, reflecting the inherent interdependence between preliminary observations and ultimate diagnostic judgments. The observed regularity implies that the dataset can decompose complex diagnostic reasoning into sequential and interpretable sub-tasks, where each step contributes complementary clinical information that collectively supports the final diagnostic interpretation. The concentration of reasoning flow within a few central nodes further supports the procedural consistency and clinical plausibility of the Step-CoT design, suggesting that it provides a structured approximation of step-by-step medical reasoning rather than a collection of isolated decision points.

\section{Enhancing Med-VQA Method with Step-CoT Dataset}
\label{sec:method_overview}
We introduce a visual CoT framework for the Step-CoT dataset, which provides stepwise supervision corresponding to clinically meaningful sub-tasks. Steps in Step-CoT have variable lengths and non-linear dependencies across diseases~\cite{scott2004lung}. To capture this, each step is modeled as a node and clinical dependencies as edges, forming a graph. A global memory node dynamically aggregates information, preserving contextual coherence and enabling interpretable multi-step reasoning.
This framework adopts a collaborative teacher-student paradigm. Further implementation details are provided in Appendix Sec. D.

\paragraph{Teacher Model Overview.} The teacher processes $S$ sequential clinical questions (steps) and an explicit \emph{memory node} that aggregates cross-step information. For each step $s\in\{1,\dots,S\}$ the model: (1) encodes the step prompt with a shared text encoder, (2) forms a node set $\{\mathbf{t}_1,\dots,\mathbf{t}_S,\mathbf{m}\}$ where $\mathbf{t}_s$ is the CLS embedding of step $s$ and $\mathbf{m}$ is the learnable memory, (3) updates node states with a multi-head Graph Attention Network (GAT), (4) composes a step context by fusing the step node and the memory node, and (5) predicts the step label using a step-specific classifier. After prediction, the teacher writes a compact prediction embedding back to the memory via a gated GRU update, enabling information flow to later steps.

\paragraph{GAT Memory.} The GAT implements multi-head attention between nodes. For a single head, after a linear map $W$, we compute attention scores:
\begin{equation}
    e_{ij} = \mathrm{LeakyReLU}\big( \mathbf{a}_{\mathrm{src}}^\top (W\mathbf{h}_i) + \mathbf{a}_{\mathrm{dst}}^\top (W\mathbf{h}_j)\big),
\end{equation}
where $\mathbf{h}_i$ and $\mathbf{h}_j$ denote the input features of node $i$ and node $j$, respectively, $W$ is a learnable linear projection matrix, and $\mathbf{a}_{\mathrm{src}}$, $\mathbf{a}_{\mathrm{dst}}$ are learnable attention vectors for source and destination nodes. The normalized attention coefficient is:
\begin{equation}
    \alpha_{ij} = \mathrm{softmax}_j(e_{ij}),
\end{equation}
where $\alpha_{ij}$ represents the attention weight from node $i$ to node $j$. The memory node corresponds to the last node in the graph and thus receives aggregated information from all step nodes.

\paragraph{Student Model and Distillation.} While the teacher graph is explicitly designed to capture rich, inter-step reasoning dependencies on our Step-CoT dataset, these learned relations are often highly structured and dataset-specific. To enable practical deployment and cross-dataset transfer, we therefore train a lightweight student model via knowledge distillation. The student aims not only to preserve the teacher’s reasoning behavior but also to learn a compressed, more generalizable representation of those complex relations. This reduces inference cost, simplifies deployment, and mitigates adaptation problems that arise when the teacher’s dataset-specific reasoning patterns are applied to datasets with different reasoning complexity or distributional characteristics.

The student model is a compact chain model that only uses image features and a sequence of light-weight heads. We distill from the teacher to the student using three complementary losses per step:

% \begin{itemize}
  % \item 
  \textbf{Hard Supervision:} cross-entropy loss on labeled examples:
  \begin{equation}
      \mathcal{L}_{\mathrm{CE}} = -\frac{1}{N}\sum_{i=1}^{N}\log p(y_i),
  \end{equation}
  where $N$ is the number of labeled examples and $p(y_i)$ is the predicted probability for the true label $y_i$.

  % \item 
  \textbf{Soft KD:} Kullback-Leibler divergence between softened teacher and student logits:
  \begin{equation}
      \mathcal{L}_{\mathrm{KD}} = T^2 \cdot \mathrm{KL}\big(\sigma(\ell_s/T)\,\|\,\sigma(\ell_t/T)\big),
  \end{equation}
  where $\ell_t$ and $\ell_s$ denote the teacher and student logits respectively, $\sigma(\cdot)$ denotes the softmax function, and $T$ is the temperature controlling softening of logits.

  % \item 
  \textbf{Channel/Relation Alignment (CH):} an HSIC-inspired~\cite{ma2020hsic} inter-example similarity alignment loss:
  \begin{equation}
      \mathcal{L}_{\mathrm{CH}} = w_{\mathrm{fw}} \cdot \mathrm{KL}\big(\log K_V \,\|\, K_U\big),
  \end{equation}
  where $K_U$ and $K_V$ are the softmax-normalized projected feature matrices of the teacher and student, respectively, and $w_{\mathrm{fw}}$ is a similarity weighting factor computed from the alignment of their centered Gram matrices, reflecting how well teacher-student feature relations align.
% \end{itemize}

\paragraph{Training Recipe.} We train with separate optimizers for teacher and student. The teacher may be optionally pre-trained for several epochs with supervised loss only, then both teacher and student are trained: the teacher receives supervised CE updates, and the student is trained to minimize
\begin{equation}
    \mathcal{L}_{\mathrm{student}}^{(s)} = \mathcal{L}_{\mathrm{CE}}^{(s)} + \alpha_{\mathrm{KD}}\mathcal{L}_{\mathrm{KD}}^{(s)} + \alpha_{\mathrm{CH}}\mathcal{L}_{\mathrm{CH}}^{(s)},
\end{equation}
where $\alpha_{\mathrm{KD}}$ and $\alpha_{\mathrm{CH}}$ are weighting coefficients that balance the contributions of soft KD and CH alignment losses.

\section{Experiments}
\subsection{Benchmark Establishment}
\begin{table}[!t]
\centering
\caption{Test results of diagnosis step on different models using Step-CoT(\%). Entries are reported as $a({+}b)$, where $a$ is the performance without Step-CoT and $b$ is the improvement with it. The best results in each column are highlighted in \textbf{bold}, and the second-best values are \underline{underlined}.}
\label{tab:step7_performance}
\resizebox{\linewidth}{!}{%
% \begin{threeparttable}
\begin{tabular}{lcccc}
\hline
Model & Accuracy & mAUC & Sensitivity & Specificity \\
\midrule
\multicolumn{5}{l}{{LVLMs}} \\ 
\midrule
LLaVA-Med~\cite{li2023llava} &42.7 &58.3 &42.7 &79.4 \\
Med-Flamingo~\cite{moor2023med} &30.1 &61.2 &28.4 &89.8 \\
\midrule
\multicolumn{5}{l}{Visual foundation models} \\
\midrule
VisualBERT~\cite{li2019visualbert} &56.2(+9.3) &48.8(+14.3) &8.5(+1.6) &89.1(+2.7) \\
CLIP~\cite{radford2021learning} &64.7(+4.5) &48.8(+3.8) &10.1(+1.9) &87.5(+2.1) \\
ALBEF~\cite{li2021align} &68.1(+3.9) &53.9(+21.2) &16.3(+2.2) &91.5(+1.8) \\
BLIP~\cite{li2022blip} &66.4(+4.6) &53.2(+21.7) &15.5(+1.7) &90.8(+2.1) \\
FLAVA~\cite{singh2022flava} &62.5(+4.6) &50.2(+14.0) &9.3(+1.6) &90.7(+1.6) \\
biomedclip~\cite{zhang2023biomedclip} &69.3(+3.8) &55.6(+20.4) &19.4(+2.3) &91.8(+1.7) \\
% \multicolumn{5}{l}{{Proposed method}} \\
% \midrule
Ours (Teacher) & \textbf{78.3} & \underline{89.5} & \textbf{46.0} & \textbf{96.6}  \\
Ours (Student) & \underline{77.5}& \textbf{90.0} & \underline{41.8} & \underline{96.0} \\
\bottomrule
\end{tabular}
}
\end{table}
We establish performance baselines using LVLMs and visual foundation models. Table~\ref{tab:step7_performance} reveals the results. 
Multi-modal models achieve good results but still exhibit sensitivity–specificity imbalance. Among them, BiomedCLIP~\cite{zhang2023biomedclip} attains the best overall performance, suggesting that domain adaptation enhances domain alignment. In contrast, LVLMs demonstrate limited transfer accuracy on this benchmark (30–40\%), reflecting the gap between generic multi-modal pretraining and domain-specific reasoning requirements. Despite their impressive open-ended generation ability, these LVLMs often rely on surface-level correlations and struggle to maintain factual precision or structured reasoning consistency across clinical steps. Regarding the Step-CoT setup, this setting specifies whether Step-CoT is enabled, allowing the model to leverage structured stepwise reasoning, or disabled, in which case the model is trained without any intermediate reasoning guidance. The results show that incorporating CoT leads to improvements across visual foundation models.
The proposed Step-CoT framework introduces stepwise supervision over intermediate reasoning processes, enabling the model to accumulate and verify evidence across diagnostic stages. This design not only improves factual precision but also leads to significant gains in interpretability.

\begin{figure}[!t]
    \centering
    \includegraphics[width=\linewidth]{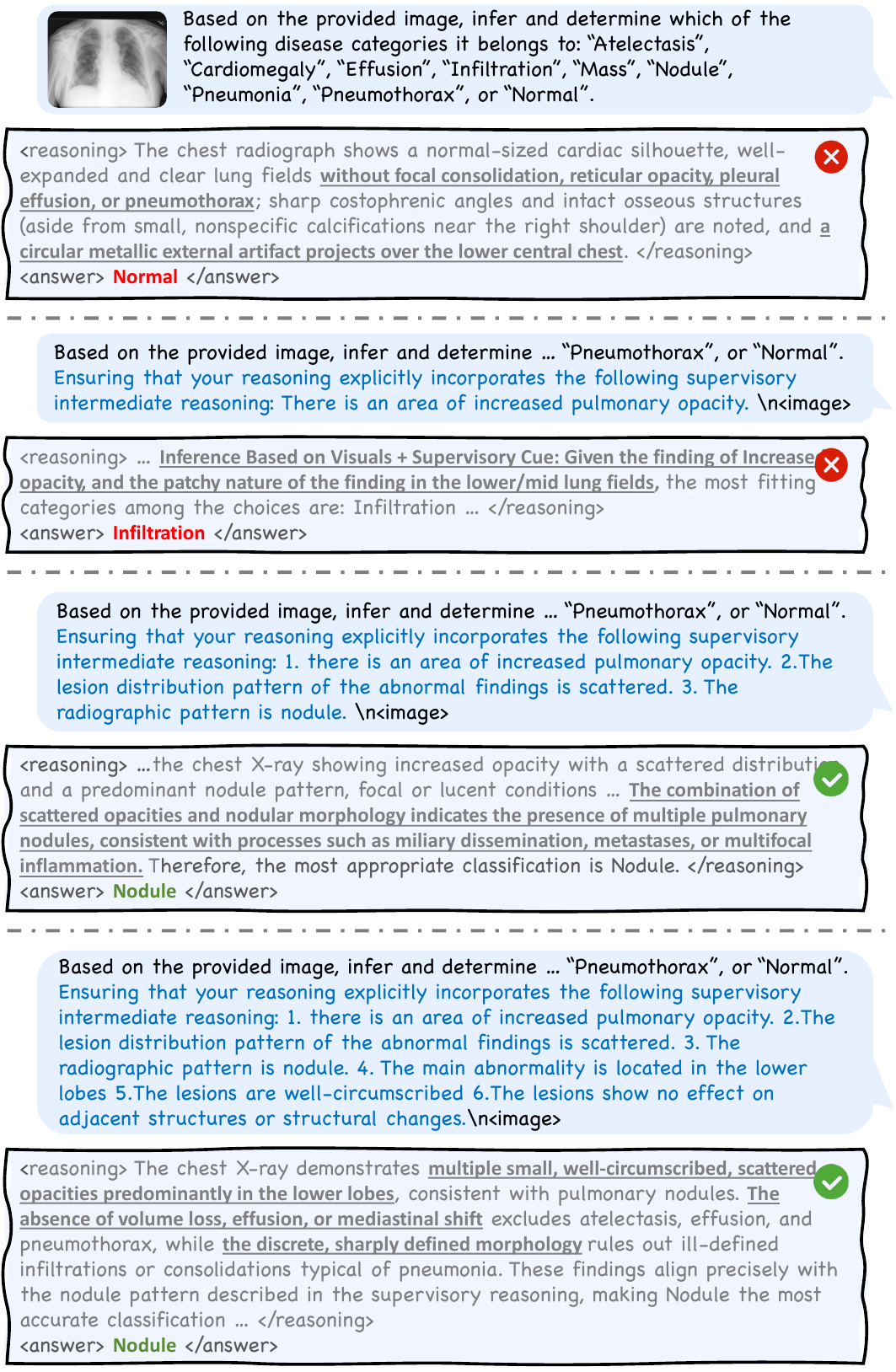}
    \caption{Fine-tuning an LVLM with Step-CoT-based intermediate constraints under verifiable instructions. The testing of different steps is conducted in independent dialogue sessions. The model progressively adjusts its stepwise reasoning, produces coherent intermediate steps, and converges to the correct final diagnosis; this demonstrates that Step-CoT’s structured intermediate constraints strengthen model reasoning and reliably guide it to accurate conclusions.}
    \label{fig:instruction}
\end{figure}

\subsection{Stepwise Effectiveness Study of the Step-CoT Dataset}
This experiment was designed to verify the effectiveness of the stepwise reasoning framework introduced in the Step-CoT dataset. To this end, we employed the verifiable instruction framework proposed in Instruction-Following Eval (IFEval)~\cite{zhou2023instruction} as an objective tool to assess whether introducing explicit step reasoning enables the model to generate more accurate final answers. Specifically, we first tasked Gemini~\cite{team2023gemini} to directly output the diagnosis and corresponding reasoning process. Subsequently, we progressively introduced intermediate reasoning constraints based on Step-CoT, requiring Gemini to base its reasoning on these provided structured constraints. Each reasoning step was governed by corresponding verifiable instructions, thereby enabling an evaluation of whether Step-CoT's intermediate reasoning constraints can enhance the model's reasoning capability and guide it toward correct conclusions.

As shown in Fig.~\ref{fig:instruction}, the baseline model often produced ill-defined or incorrect predictions, whereas the model guided by Step-CoT reasoning exhibited coherent intermediate reasoning and correctly localized and classified the lesions. Quantitative evaluation under the same verification framework further demonstrated that the step-supervised model achieved higher accuracy across multiple reasoning categories. These findings confirm that the stepwise reasoning mechanism embedded in Step-CoT effectively enhances model reasoning reliability and diagnostic accuracy, providing empirical evidence for the value of structured, verifiable reasoning in medical visual question answering.

\subsection{Cross-Dataset Generalization Evaluation}
To evaluate the generalization capability and robustness of models trained on our Step-CoT dataset, we conduct a cross-dataset evaluation. Following the training on Step-CoT, models are directly evaluated on the ChestX-ray8 dataset~\cite{wang2017chestx} without any further fine-tuning. This rigorous assessment aims to verify whether the multi-step reasoning skills acquired from Step-CoT can transfer effectively to a different clinical benchmark. The ChestX-ray8~\cite{wang2017chestx}, released by the U.S. National Institutes of Health (NIH), is a large-scale chest X-ray image database containing over 100k images from tens of thousands of patients. It is annotated with common thoracic diseases such as atelectasis, cardiomegaly, and effusion. In this study, we utilized the single-label classification subsets defined in the original dataset as our research subjects. It contains 985 single-labeled samples belonging to the following categories: atelectasis, cardiomegaly, effusion, infiltration, mass, nodule, pneumonia, and pneumothorax.

\begin{table}[!t]
\centering
\caption{Transfer performance comparison with and without using Step-CoT on a different test dataset (\%). Entries are reported as $a({+}b)$, where $a$ is the performance without Step-CoT and $b$ is the improvement with it. The best value in each column is \textbf{bold} and the second best is \underline{underlined}.}
% “+” indicates results obtained using Step-CoT training, while results without “+” denote models trained only with final diagnostic supervision.
\label{tab:tranfer}
\setlength{\tabcolsep}{7pt} % 控制列间距
\resizebox{0.5\textwidth}{!}{%
\begin{tabular}{lcccc}
\hline
Model & Accuracy & mAUC & Sensitivity & Specificity \\
\toprule
VisualBERT~\cite{li2019visualbert} & 25.7(+6.3) & 52.1(+4.3) & 30.8(+4.7) & 73.9(+3.9) \\
CLIP~\cite{radford2021learning} & 26.8(+7.3) & 53.3(+4.6) & 18.4(+3.3) & 79.7(+3.7) \\
ALBEF~\cite{li2021align} & 30.4(+5.2) & \underline{58.3(+3.1)} & \underline{38.4(+3.5)} & 81.5(+2.9) \\
BLIP~\cite{li2022blip} & 27.5(+6.1) & 52.7(+3.7) & 17.9(+2.9) & 78.2(+3.5) \\
FLAVA~\cite{singh2022flava} & 28.9(+5.5) & 56.8(+2.9) & 35.5(+4.1) & 77.3(+3.0) \\
biomedclip~\cite{zhang2023biomedclip} & \underline{33.6(+5.8)} & \textbf{61.9(+3.6)} & \textbf{42.7(+3.9)} & \textbf{85.6(+2.6)} \\

Ours (Teacher) & \textbf{40.3} & 61.3 & 20.2 & \underline{86.1}  \\
Ours (Student) & 39.1 & 57.4 & 23.5 & 80.7 \\
\bottomrule
\end{tabular}}
\end{table}

The results in Table~\ref{tab:tranfer} demonstrate that networks trained with Step-CoT supervision consistently outperform their non-step counterparts when transferred to the ChestX-ray8 benchmark. The results indicate that the Step-CoT training regime improves both discrimination and clinical relevance after cross-dataset transfer. The fact that both the distilled Student and the Teacher, each trained under the Step-CoT paradigm, perform competitively on ChestX-ray8 suggests that the stepwise knowledge is transferable through distillation and that the learned stepwise reasoning is robust to dataset shift. These results support that Step-CoT enables the model to learn structured, stepwise diagnostic logic rather than relying on end-to-end correlations.

\subsection{Ablation Experiments}
We evaluate the proposed GAT-based teacher model with knowledge distillation against established baselines.

\begin{table}[!t]
\centering
\caption{Comparison of clinical expert evaluation across four of seven reasoning steps (\%). The best results in each column are highlighted in \textbf{bold}, and the second-best values are \underline{underlined}.}
\label{tab:clinician}
\setlength{\tabcolsep}{10pt} % 控制列间距
\resizebox{\linewidth}{!}{%
\begin{threeparttable}
\begin{tabular}{lcccc}
\toprule
 & Detection & Distribution & Location & Diagnosis\\
\midrule
Clinician\tnote{$\dagger$} & 72.1 & 66.0 & 66.0 & \underline{73.1} \\
Teacher\tnote{$\dagger$} & \textbf{88.5} & \textbf{78.4} & \textbf{72.8} & \textbf{79.8} \\
Student\tnote{$\dagger$} & \underline{80.4} & \underline{72.6} & \underline{69.5} & 68.5 \\
\bottomrule
\end{tabular}
\begin{tablenotes}\footnotesize
\item[{$\dagger$}] Test sample size: 200 cases.
\end{tablenotes} 
\end{threeparttable}%
}
\end{table}

\begin{table}[!t]
\centering
\caption{Comparison of memory ablation results across four of seven reasoning steps (\%). The best results in each column are highlighted in \textbf{bold}, and the second-best values are \underline{underlined}.}
\label{tab:memory}
\setlength{\tabcolsep}{10pt} % 控制列间距
\resizebox{\linewidth}{!}{%
\begin{tabular}{lcccc}
\toprule
 & Detection & Distribution & Location & Diagnosis\\
\midrule
w/o Memory & 73.7 & 69.6 & 63.2 & 65.5 \\
w/o Text & \underline{81.5} & {76.1} & {69.3} & {72.1} \\
Ours (Teacher) & \textbf{91.8} & \textbf{84.6} & \textbf{77.1} & \textbf{78.3} \\
Ours (Student) & \textbf{91.8} & \underline{83.4} & \underline{76.9} & \underline{77.5} \\
\bottomrule
\end{tabular}
}
\end{table}

\textbf{Comparative Performance Analysis.} To evaluate the proposed GAT-Memory framework, we compare it with two ablated variants. The \emph{w/o Memory} variant removes the memory module and GRU update, disabling cross-step information accumulation and thus breaking the continuity of multi-step reasoning. The \emph{w/o Text} variant omits textual prompts, relying only on visual features. As reported in Table~\ref{tab:memory}, both ablations cause consistent performance drops across steps: removing the memory produces the largest accuracy decline (65.45\%), highlighting the necessity of temporal state propagation for synthesizing intermediate evidence, while excluding textual prompts also reduces performance (72.10\%), confirming the role of linguistic guidance in grounding visual interpretation. The full Teacher model attains the highest accuracy (78.26\%), and the distilled Student closely matches this performance (77.53\%) with reduced complexity. Taken together, these findings highlight key insights: (i) the memory module is indispensable for maintaining contextual reasoning across diagnostic steps, enabling the model to “think with images” in a clinically coherent manner; (ii) the student model effectively inherits the teacher’s structured diagnostic logic while achieving lightweight, transferable inference.

\textbf{Clinical Expert Evaluation}. We evaluated 200 randomly selected cases. For each case, we compared three outputs: Clinician (expert), Teacher (model), and Student (model), across four phased diagnostic steps. As shown in Table~\ref{tab:clinician}, the Teacher model consistently outperforms both the Student and clinician baselines across all diagnostic steps. The Student model closely follows, maintaining performance within 5-8\% of the Teacher on most steps. Notably, both models surpass clinician-level accuracy in mid-level reasoning tasks such as Distribution and Location, suggesting that stepwise supervision enables more consistent and fine-grained feature interpretation. These results confirm that the Step-CoT framework effectively captures clinically coherent reasoning patterns and that the distilled Student model preserves this structured diagnostic competence with minimal performance loss. Detailed per-step accuracies are provided in Appendix Sec C.

\begin{figure}
    \centering
    \includegraphics[width=\linewidth, trim=0cm 0cm 0.25cm 0cm, clip]{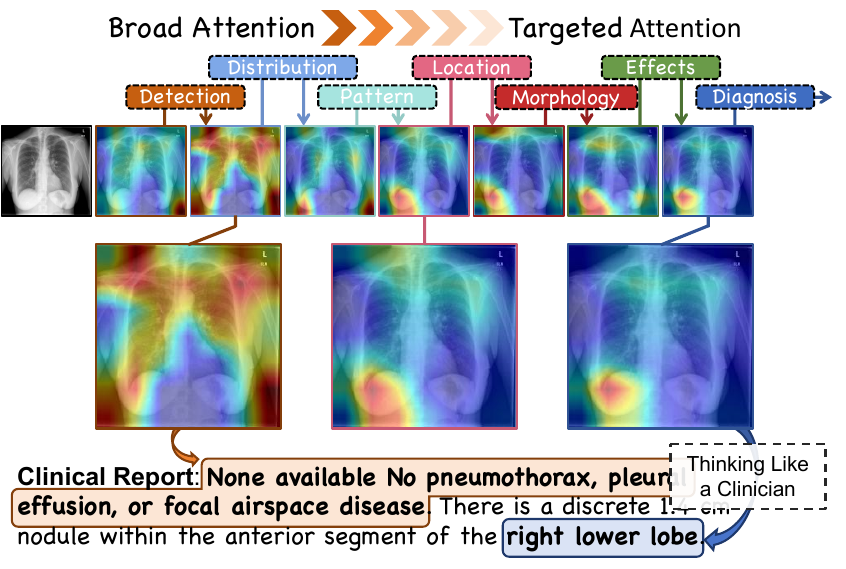}
    \caption{The feature attention visualization across multi-step reasoning demonstrates an evolution from broad attention in the initial query steps to highly targeted attention in the final diagnostic step, reflecting the multi-step capability of Step-CoT and visually verifying the effectiveness of the reasoning chain.
    }
    \label{fig:visualization}
\end{figure}
%%%%%%%%%%%%%%%%%%%%%%%%%
\subsection{Visualization of the Reasoning Steps}
Fig.~\ref{fig:visualization} presents the seven stepwise attention maps, which visualize how the model’s focus evolves during reasoning. Across the sequence, the attention progressively concentrates from broad, image-level saliency to fine-grained, lesion-specific regions: early steps highlight global abnormality and distribution patterns, middle steps emphasize modality-specific imaging cues and precise lesion localization, and the final steps concentrate on diagnostic features.

This ordered sharpening of attention provides direct, interpretable evidence that the model engages in image-guided, multi-step reasoning rather than a single opaque decision. The maps serve two roles: (i) they demonstrate the model’s emerging multi-step capability by showing how intermediate visual evidence is assembled into later diagnostic judgments; (ii) they make the diagnostic chain traceable, enabling qualitative inspection of which image regions and which reasoning stages contribute to the final prediction.

\section{Conclusion}
In this work, we presented Step-CoT, a large-scale, clinically grounded medical reasoning dataset designed to bring multi-step reasoning into Med-VQA. By explicitly supervising intermediate reasoning with expert-curated diagnostic steps, Step-CoT enables models to think with multi-step, progressively aligning visual attention and linguistic inference to follow clinically valid diagnostic pathways, enhancing interpretability while maintaining diagnostic precision. Building upon this foundation, our teacher-student CoT framework effectively learn from Step-CoT, enhancing the efficiency of multi-step reasoning. Extensive experiments confirm that Step-CoT establishes a structured and credible paradigm for medical reasoning, bridging the gap between human clinical cognition and AI-based decision-making. We expect Step-CoT to serve as a cornerstone resource for developing next-generation, trustworthy medical VQA systems.

{
    \small
    \bibliographystyle{ieeenat_fullname}
    \bibliography{main}
}

\clearpage
\setcounter{page}{1}
\maketitlesupplementary
\appendix
%%%%%%%%%%%%%%%%%%%%%%%%%%%%%%%%%%%%%%%%%%%%%%%%%%%%%%%%%%%%%%%%
\section{Step-CoT Data Access and Format}
The data can be accessed on HuggingFace at... The benchmark and code can be accessed on GitHub at...
The dataset is organised in one main folder corresponding to three datasets. The dataset structure is shown as follows:
\begin{tcolorbox}[
    colback=gray!10,      % 灰色背景
    colframe=gray!50,     % 灰色边框
    sharp corners,
    boxsep=3pt,
    left=6pt,
    right=6pt,
    top=6pt,
    bottom=6pt
]
\begin{verbatim}
Step-CoT/
|-- train.xlsx    # train paths
|-- val.xlsx      # val paths
|-- test.xlsx     # test paths
|
|-- dataset/
|   |-- data.json
|   |-- images/
|       |-- 1_IM-0001-4001.dcm.png
|       |-- 2_IM-0652-1001.dcm.png
|       |-- 100..._370g35.png
|       |-- 100..._ydw2jy.png
|       |-- 1_1.png
|       |-- 2_1.png
\end{verbatim}
\end{tcolorbox}
\begin{enumerate}
    \item StepCoT/dataset/images: Contains the frontal-view original CXR images from all datasets. The image filenames follow the naming conventions of their respective source datasets.
    \item StepCoT/dataset/data.json: Includes the stepwise VQA questions, answers, and associated reasoning for each image. The format of entries in the JSON file is shown as follows:
    \begin{tcolorbox}[title=data.json, breakable]
\begin{lstlisting}[language=json]
  {
    "patient_id": "***",
    "image_path": "***",
    "origin": "***",
    "report": "The cardiac silhouette and mediastinum size are within normal limits. There is no pulmonary edema. There is no focal consolidation. There are no signs of a pleural effusion. There is no evidence of pneumothorax.",
    "vqa_chain": [
      {
        "step": "Step 1",
        "question": "Is there any abnormal radiodensity in the lungs?",
        "options": [
          "No abnormality",
          "A1. Increased opacity",
          "A2. Decreased opacity",
          "A3. Mixed"
        ],
        "answer": "No abnormality",
        "reasoning": "The report explicitly states no focal consolidation or other lung abnormalities, and the impression confirms a normal chest x-ray."
      },
      {
        "step": "Step 2",
        "question": "What is the distribution pattern of the abnormal findings?",
        "options": [
          "B1. Focal",
          "B2. Scattered",
          "B3. Diffuse"
        ],
        "answer": "N/A",
        "reasoning": "No abnormalities exist, making distribution patterns irrelevant."
      },
      {
        "step": "Step 3",
        "question": "What is the predominant imaging pattern?",
        "options": [
          "C1. Consolidation",
          "C2. Nodule"
          "C3. Reticular",
          "C4. Cavity",
          "C5. Ground-glass opacity",
        ],
        "answer": "N/A",
        "reasoning": "No abnormalities exist, making imaging patterns irrelevant."
      },
      {
        "step": "Step 4",
        "question": "Where is the main abnormality located?",
        "options": [
          "D1. Pleura/Chest wall",
          "D2. Lower lobes",
          "D3. Bilateral diffuse",
          "D4. Left upper lobe",
          "D5. Right upper lobe",
          "D6. Mediastinum"
        ],
        "answer": "N/A",
        "reasoning": "No abnormalities exist, making location irrelevant."
      }
    {
        "step": "Step 5",
        "question": "Are the lesions well-defined or have any internal characteristics?",
        "options": [
          "E1. Scarring/Fibrosis"
          "E2. Well-circumscribed",
          "E3. Spiculated",
          "E4. Cavitary",
        ],
        "answer": "N/A",
        "reasoning": "No abnormalities exist, making lesion characteristics irrelevant."
      },
      {
        "step": "Step 6",
        "question": "Do the lesions affect adjacent structures or cause structural changes?",
        "options": [
          "F1. Hyperinflation"
          "F2. Pleural effusion",
          "F3. Volume loss/atelectasis",
          "F4. No effect",
          "F5. Pneumothorax",
          "F6. Mediastinal shift",
        ],
        "answer": "N/A",
        "reasoning": "No abnormalities exist, making structural effects irrelevant."
      },
      {
        "step": "Step 7",
        "question": "What is the most likely radiographic diagnosis?",
        "options": [
          "1. Normal"
          "2. Infiltration",
          "3. Effusion",
          "4. Pneumonia",
          "5. Atelectasis",
          "6. Nodule",
          "7. Cardiomegaly",
          "8. Mass",
          "9. Pneumothorax", 
        ],
        "answer": "Normal",
        "reasoning": "The impression concludes a normal chest x-ray with no acute findings."
      }
    ]
  }
\end{lstlisting}
\end{tcolorbox}
The key columns are described as follows:
\begin{itemize}
    \item \textbf{patient\_id}: Each anonymous patient identifier corresponds to a single sample, and each sample contains only one anteroposterior (frontal) chest X-ray image. These identifiers ensure subject anonymity while allowing each CXR instance to be uniquely and consistently tracked throughout the dataset.
    \item \textbf{image\_path}: A file reference pointing to the corresponding radiographic image for each sample. This field provides the exact storage location of the CXR image within the dataset directory structure, enabling reliable retrieval and consistent linkage between metadata entries and their associated medical images.
    \item \textbf{origin}: Dataset source information indicating the original dataset from which each sample was collected, ensuring traceability across heterogeneous data sources and enabling proper dataset-level stratification or analysis when required.
    \item \textbf{report}: Original radiology report text containing the clinician’s narrative description of the image, including lesion characteristics, anatomical location, and other relevant diagnostic observations. All patient-identifiable or sensitive personal information has been fully removed to ensure compliance with privacy.
    \item \textbf{vqa\_chain}: Seven-step diagnostic reasoning sequence:
    \begin{enumerate}
        \item Detection step
        \item Lesion distribution step
        \item Radiographic pattern step
        \item Anatomical location step
        \item Morphologic feature step
        \item Secondary effects/associated signs step
        \item Diagnosis step
    \end{enumerate}
\end{itemize}

Each VQA step contains the question, multiple-choice options, selected answer, and clinical reasoning, creating a comprehensive framework for structured radiological interpretation and AI model training.
\end{enumerate}
%%%%%%%%%%%%%%%%%%%%%%%%%%%%%%%%%%%%%%%%%%%%%%%%%%%%%%%%%%%%%%%%
\section{Detailed Information of Step-CoT}
\subsection{Data Acquisition}
This work comprises original frontal CXR images and associated diagnostic text drawn from three public sources (totaling 10,068 CXR studies): (i) IU X-Ray \cite{demner2016preparing}, from which we use a subset of 3,749 CXR studies; (ii) PadChest-GR \cite{castro2024padchest}, from which we use a subset of 3,230 CXR studies; and (iii) Med-Image-Reports \footnote{https://huggingface.co/datasets/zirui3/med-image-reports}, from which we use a subset of 3,089 CXR studies. The combined corpus enables experiments on image–report alignment, grounded report generation, and stepwise VQA supervision across a broad mix of normal and abnormal cases.

% \begin{table}[ht]
%   \centering
%   \caption{Datasets and subset sizes used in Step-CoT.}
%   \vspace{3pt}
%   \begin{tabular}{l r}
%     \hline
%     Dataset & Number of CXR studies used \\
%     \hline
%     IU X-Ray \cite{demner2016preparing} & 3,749 \\
%     PadChest-GR \cite{castro2024padchest} & 3,230 \\
%     Med-Image-Reports\footnote{\url{https://huggingface.co/datasets/zirui3/med-image-reports}} & 3,089 \\
%     \hline
%     {Total} & {10,068} \\
%     \hline
%   \end{tabular}
%   \label{tab:datasets_overview}
% \end{table}

\subsubsection{IU X-Ray}
The IU X-Ray dataset was collected by Indiana University and contains a large corpus of chest radiographs with associated radiology reports. Reports are organized under several headings (``Findings'', ``Impression'', ``Comparison'', and ``Indication''); for this study, we use captions from the ``Findings'' section to provide descriptive image-level text. From the available corpus, we selected a curated subset of 3,749 frontal CXR studies that meet our inclusion criteria.

\subsubsection{PadChest-GR}
PadChest-GR is a bilingual chest X-ray benchmark derived from PadChest and tailored for Grounded Radiology Report Generation. The dataset includes clinician-validated annotations, bounding-box grounding for findings, and structured metadata; reports were processed (including sentence extraction, English translation, and label linking) to produce high-quality, sentence-level finding annotations. A team of radiologists further refined the corpus by removing low-quality studies and annotating bounding boxes and categorical labels. For our experiments, we use a subset of 3,230 frontal-view CXR studies drawn from the PadChest-GR release.

\subsubsection{Med-Image-Reports}
The Med-Image-Reports benchmark aggregates chest X-ray studies and radiology-style captions from multiple public sources (OpenI, MIMIC-CXR, and PadChest). Original reports were preprocessed into concise diagnostic-style captions that describe both normal structures and clinically relevant abnormalities (e.g., cardiomegaly, pulmonary opacity, pleural effusion, pneumothorax, presence of devices). We adopt these standardized captions to ensure consistent supervision across heterogeneous origins and use a subset of 3,089 CXR studies from the Med-Image-Reports collection.

\paragraph{Preprocessing and harmonization}
Across all three sources, we restrict to frontal-view studies, extract or select the diagnostic caption, normalize common clinical terms, and remove records with missing or unusable captions. After harmonization, the resulting corpus contains 10,068 CXR studies used throughout the experiments reported in this paper.

\subsection{Data Annotation}
To construct a unified stepwise VQA supervision protocol across heterogeneous chest X-ray datasets, we perform automated annotation based on the corresponding radiology reports. For each CXR study, we utilize a large-scale language model (DeepSeek) to parse the paired report and extract clinically grounded information aligned with our seven-step diagnostic reasoning framework. Specifically, the model is prompted to identify key radiological observations, synthesize diagnostic cues, and populate each step of the VQA schema with structured outputs, including the step-specific question, the corresponding answer, and a concise reasoning explanation. This automated annotation pipeline ensures consistent interpretation across datasets while preserving the clinical semantics embedded in expert-written reports. The complete prompt used for generating Step-CoT annotations is provided below.
\begin{RoundedTitleBox}{LLM Prompt}
Please analyze the chest X-ray findings based on the provided patient report (including patient\_id, image\_path, report, etc. (Replace according to the information contained in different datasets.) The output must strictly adhere to JSON format, containing patient\_id, image\_path, report, and vqa\_chain. vqa\_chain is an array where each element represents a step, including step, question, options, answer, and reasoning.

Answer rules:
1. Use "No Answer" if the report does not mention the relevant answer.
2. Use "N/A" if there is no tumor or the question is not applicable (e.g., when Step 1 answer indicates no abnormality, subsequent questions about tumors should be "N/A").
3. For "No Answer" cases, attempt to infer the answer based on information from label, locations, sentence\_en, label\_group, classification, etc., if possible; otherwise, maintain "No Answer".
4. When providing answers, must consider previous steps' answers to establish a chain of thought.

Reasoning rules:
1. Must be written in English as complete sentences.
2. Based on report content or logical inference, directly analyze the reasons without additional modifiers (e.g., "the report describes").
3. In reasoning, must consider previous steps' answers to establish a chain of thought (e.g., referencing Step 1 conclusions to explain Step 2 answers).

Format rules:
1. step, question, and options in vqa\_chain must be directly copied from the following template to ensure textual consistency.
2. The entire output should be a JSON object.

Template vqa\_chain structure:
[
  {"step": "Step 1", "question": "Is there any abnormal radiodensity in the lungs?", "options": ["No abnormality", "Increased opacity", "Decreased opacity", "Mixed"]},
  {"step": "Step 2", "question": "What is the distribution pattern of the abnormal findings?", "options": ["Focal", "Scattered", "Diffuse"]},
  {"step": "Step 3", "question": "What is the predominant imaging pattern?", "options": ["Consolidation", "Ground-glass opacity", "Reticular", "Cavity", "Nodule"]},
  {"step": "Step 4", "question": "Where is the main abnormality located?", "options": ["Right upper lobe", "Left upper lobe", "Lower lobes", "Bilateral diffuse", "Pleura/Chest wall", "Mediastinum"]},
  {"step": "Step 5", "question": "Are the lesions well-defined or have any internal characteristics?", "options": ["Well-circumscribed", "Spiculated", "Cavitary", "Scarring/Fibrosis"]},
  {"step": "Step 6", "question": "Do the lesions affect adjacent structures or cause structural changes?", "options": ["No effect", "Mediastinal shift", "Volume loss/atelectasis", "Pleural effusion", "Pneumothorax", "Hyperinflation"]},
  {"step": "Step 7", "question": "What is the most likely radiographic diagnosis?", "options": ["Atelectasis", "Cardiomegaly", "Effusion", "Infiltration", "Mass", "Nodule", "G. Pneumonia", "Pneumothorax", "Normal"]}
]
\end{RoundedTitleBox}

\subsubsection{LLM Prompt Consistency Experiment}
We conduct an evaluation to quantify how different LLMs respond to an identical prompt for generating stepwise VQA annotations from chest X-ray reports. The same prompt and input set are submitted, verbatim, to three representative models (DeepSeek, ChatGPT, and G-Mini) under fixed decoding settings. For each report, we collect the structured JSON outputs and present side-by-side comparisons of the vqa\_chain entries; the analysis focuses on per-step categorical agreement as well as differences in the free-text reasoning. This experimental design isolates prompt-driven variance by holding inputs and decoding parameters constant and provides direct, empirical evidence of how different LLMs interpret identical clinical text.
\begin{tcolorbox}[title=DeepSeek output,colback=gray!4,breakable]
\small
\begin{lstlisting}[breaklines=true,breakatwhitespace=false,columns=fullflexible]

{
 "patient_id": "1",
 "image_path": "",
 "report": "The cardiac silhouette and mediastinum size are within normal limits. There is no pulmonary edema. There is no focal consolidation. There are no signs of a pleural effusion. There is no evidence of pneumothorax.",
 "vqa_chain": [
  {
   "step": "Step 1",
   "question": "Is there any abnormal radiodensity in the lungs?",
   "options": ["No abnormality", "Increased opacity", "Decreased opacity", "Mixed"],
   "answer": "No abnormality",
   "reasoning": "The report explicitly states no focal consolidation or other lung abnormalities, and the impression confirms a normal chest x-ray."
  },
  {
   "step": "Step 2",
   "question": "What is the distribution pattern of the abnormal findings?",
   "options": ["Focal", "Scattered", "Diffuse"],
   "answer": "N/A",
   "reasoning": "No abnormalities exist, making distribution patterns irrelevant."
  },
  {
   "step": "Step 3",
   "question": "What is the predominant imaging pattern?",
   "options": ["Consolidation", "Ground-glass opacity", "Reticular", "Cavity", "Nodule"],
   "answer": "N/A",
   "reasoning": "No abnormalities exist, making imaging patterns irrelevant."
  },
  {
   "step": "Step 4",
   "question": "Where is the main abnormality located?",
   "options": ["Right upper lobe", "Left upper lobe", "Lower lobes", "Bilateral diffuse", "Pleura/Chest wall", "Mediastinum"],
   "answer": "N/A",
   "reasoning": "No abnormalities exist, making location irrelevant."
  },
  {
   "step": "Step 5",
   "question": "Are the lesions well-defined or have any internal characteristics?",
   "options": ["Well-circumscribed", "Spiculated", "Cavitary", "Scarring/Fibrosis"],
   "answer": "N/A",
   "reasoning": "No abnormalities exist, making lesion characteristics irrelevant."
  },
  {
   "step": "Step 6",
   "question": "Do the lesions affect adjacent structures or cause structural changes?",
   "options": ["No effect", "Mediastinal shift", "Volume loss/atelectasis", "Pleural effusion", "Pneumothorax", "Hyperinflation"],
   "answer": "N/A",
   "reasoning": "No abnormalities exist, making structural effects irrelevant."
  },
  {
   "step": "Step 7",
   "question": "What is the most likely radiographic diagnosis?",
   "options": ["Atelectasis", "Cardiomegaly", "Effusion", "Infiltration", "Mass", "Nodule", "Pneumonia", "Pneumothorax", "Normal"],
   "answer": "Normal",
   "reasoning": "The impression concludes a normal chest x-ray with no acute findings."
  }
 ]
}
\end{lstlisting}
\end{tcolorbox}

\begin{tcolorbox}[title=ChatGPT output (example),colback=gray!4,breakable]
\small
\begin{lstlisting}[breaklines=true,breakatwhitespace=false,columns=fullflexible]

{
 "patient_id": "1",
 "image_path": "",
 "report": "The cardiac silhouette and mediastinum size are within normal limits. There is no pulmonary edema. There is no focal consolidation. There are no signs of a pleural effusion. There is no evidence of pneumothorax.",
 "vqa_chain": [
  {
   "step": "Step 1",
   "question": "Is there any abnormal radiodensity in the lungs?",
   "options": ["No abnormality", "Increased opacity", "Decreased opacity", "Mixed"],
   "answer": "A",
   "reasoning": "The report repeatedly negates pulmonary edema and focal consolidation; these explicit negative findings indicate no abnormal lung radiodensity on the frontal CXR."
  },
  {
   "step": "Step 2",
   "question": "What is the distribution pattern of the abnormal findings?",
   "options": ["Focal", "Scattered", "Diffuse"],
   "answer": "N/A",
   "reasoning": "Given the absence of pathological findings in Step 1, distribution is not applicable for this case."
  },
  {
   "step": "Step 3",
   "question": "What is the predominant imaging pattern?",
   "options": ["Consolidation", "Ground-glass opacity", "Reticular", "Cavity", "Nodule"],
   "answer": "N/A",
   "reasoning": "No opacities or abnormal imaging patterns are described; therefore, pattern classification cannot be assigned."
  },
  {
   "step": "Step 4",
   "question": "Where is the main abnormality located?",
   "options": ["Right upper lobe", "Left upper lobe", "Lower lobes", "Bilateral diffuse", "Pleura/Chest wall", "Mediastinum"],
   "answer": "N/A",
   "reasoning": "No focal abnormality was detected in earlier steps, so localization is not applicable."
  },
  {
   "step": "Step 5",
   "question": "Are the lesions well-defined or have any internal characteristics?",
   "options": ["Well-circumscribed", "Spiculated", "Cavitary", "Scarring/Fibrosis"],
   "answer": "N/A",
   "reasoning": "There are no lesions reported; lesion characterization is therefore not relevant."
  },
  {
   "step": "Step 6",
   "question": "Do the lesions affect adjacent structures or cause structural changes?",
   "options": ["No effect", "Mediastinal shift", "Volume loss/atelectasis", "Pleural effusion", "Pneumothorax", "Hyperinflation"],
   "answer": "N/A",
   "reasoning": "Absence of abnormalities implies no structural effects such as effusion or pneumothorax."
  },
  {
   "step": "Step 7",
   "question": "What is the most likely radiographic diagnosis?",
   "options": ["Atelectasis", "Cardiomegaly", "Effusion", "Infiltration", "Mass", "Nodule", "Pneumonia", "Pneumothorax", "Normal"],
   "answer": "Normal",
   "reasoning": "The combined negative findings and normal cardiac/mediastinal contours support a final impression of a normal chest radiograph."
  }
 ]
}
\end{lstlisting}
\end{tcolorbox}

\begin{figure*}
    \centering
    \includegraphics[width=\linewidth]{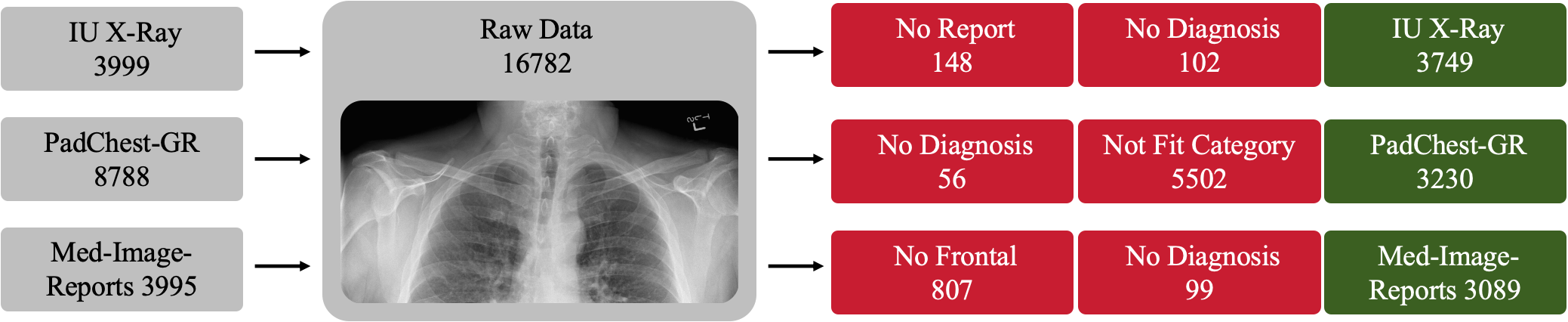}
    \caption{This study collected a total of 16,782 CXR samples in PNG format from three datasets, containing 3,999, 8,788, and 3,995 samples, respectively. After filtering, 10,068 samples were retained, yielding 10,068*7 QA pairs for training the stepwise Med-VQA task.}
    \label{fig:pre}
\end{figure*}
\begin{figure*}
    \centering
    \includegraphics[width=\linewidth]{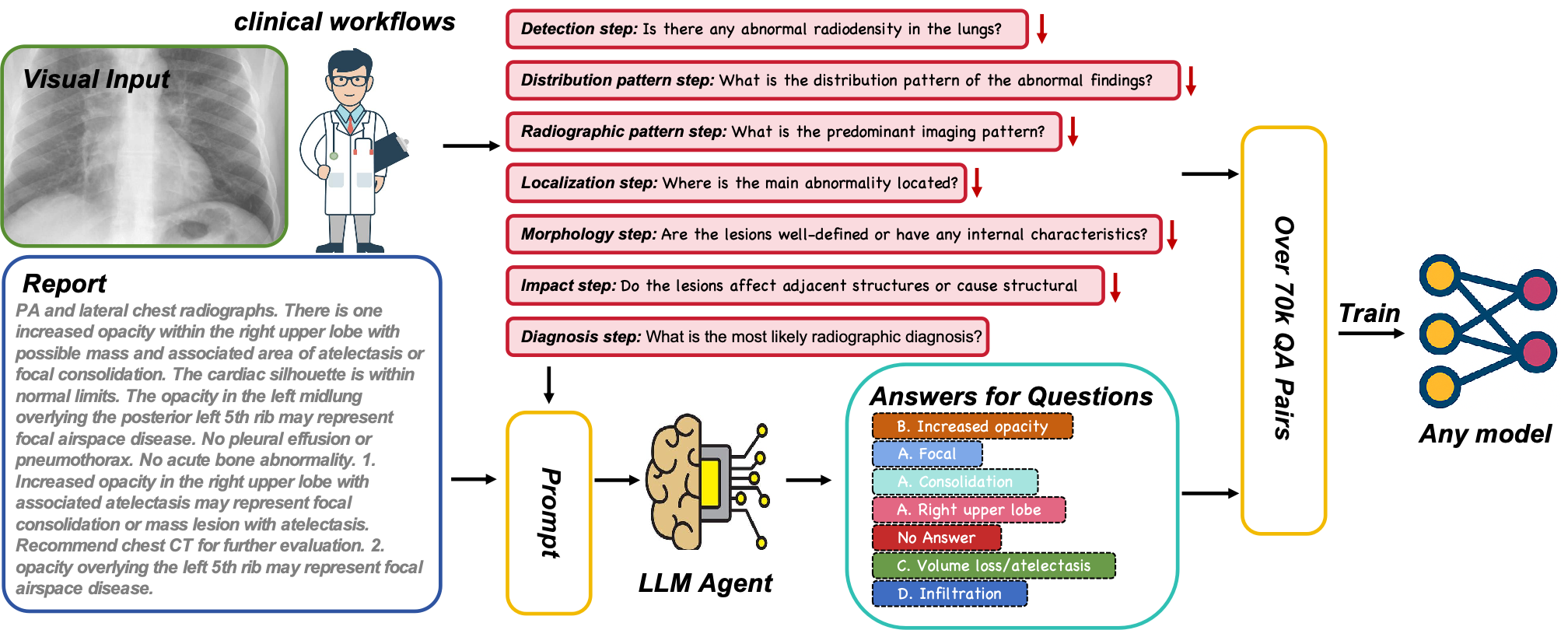}
    \caption{Preprocessing pipeline of the Step-CoT dataset.}
    \label{fig:generation}
\end{figure*}

\begin{tcolorbox}[title=Gemini output (example),colback=gray!4,breakable]
\small
\begin{lstlisting}[breaklines=true,breakatwhitespace=false,columns=fullflexible]
{
 "patient_id": "1",
 "image_path": "",
 "report": "The cardiac silhouette and mediastinum size are within normal limits. There is no pulmonary edema. There is no focal consolidation. There are no signs of a pleural effusion. There is no evidence of pneumothorax.",
 "vqa_chain": [
  {
   "step": "Step 1",
   "question": "Is there any abnormal radiodensity in the lungs?",
   "options": ["No abnormality", "Increased opacity", "Decreased opacity", "Mixed"],
   "answer": "No abnormality",
   "reasoning": "The report explicitly excludes edema, consolidation, effusion, and pneumothorax; therefore, no abnormal pulmonary radiodensity is present."
  },
  {
   "step": "Step 2",
   "question": "What is the distribution pattern of the abnormal findings?",
   "options": ["Focal", "Scattered", "Diffuse"],
   "answer": "N/A",
   "reasoning": "No abnormalities are identified at Step 1, so distribution is not applicable."
  },
  {
   "step": "Step 3",
   "question": "What is the predominant imaging pattern?",
   "options": ["Consolidation", "Ground-glass opacity", "Reticular", "Cavity", "Nodule"],
   "answer": "N/A",
   "reasoning": "Because no pathological patterns are described in the report, imaging pattern selection is not relevant."
  },
  {
   "step": "Step 4",
   "question": "Where is the main abnormality located?",
   "options": ["Right upper lobe", "Left upper lobe", "Lower lobes", "Bilateral diffuse", "Pleura/Chest wall", "Mediastinum"],
   "answer": "N/A",
   "reasoning": "There is no focal abnormality to localize given the normal findings above."
  },
  {
   "step": "Step 5",
   "question": "Are the lesions well-defined or have any internal characteristics?",
   "options": ["Well-circumscribed", "Spiculated", "Cavitary", "Scarring/Fibrosis"],
   "answer": "N/A",
   "reasoning": "No lesions are reported; internal lesion characterization cannot be performed."
  },
  {
   "step": "Step 6",
   "question": "Do the lesions affect adjacent structures or cause structural changes?",
   "options": ["No effect", "Mediastinal shift", "Volume loss/atelectasis", "Pleural effusion", "Pneumothorax", "Hyperinflation"],
   "answer": "N/A",
   "reasoning": "Given the absence of radiographic abnormalities, there is no evidence of structural impact."
  },
  {
   "step": "Step 7",
   "question": "What is the most likely radiographic diagnosis?",
   "options": ["Atelectasis", "Cardiomegaly", "Effusion", "Infiltration", "Mass", "Nodule", "Pneumonia", "Pneumothorax", "Normal"],
   "answer": "Normal",
   "reasoning": "All reported findings are negative for acute pathology and cardiac/mediastinal contours are normal, yielding a radiographic impression of normal."
  }
 ]
}
\end{lstlisting}
\end{tcolorbox}

% ---- Short comparative analysis (add after the boxes) ----

\paragraph{Comparative analysis of model outputs.}
All three models produce identical categorical answers for every step in this sample, indicating perfect per-step agreement on this case. Differences are confined to the free-text \textit{reasoning}, which are stylistic and mildly variable in focus:

\begin{itemize}
  \item {DeepSeek (used in this study)}: concise, directly references the explicit negative findings; reasoning is compact and clinically focused.
  \item {ChatGPT}: slightly more formal and explicit about cross-step justification.
  \item {Gemini}: more narrative and slightly more verbose, highlighting report completeness.
\end{itemize}

Overall, the inter-model discrepancy for this case is minimal and largely stylistic. For downstream uses that depend only on categorical answers (the vqa\_chain labels), the three models are functionally equivalent on this example. When evaluating the quality of the free-text reasoning, DeepSeek appears marginally preferable here because its explanations are concise and tightly coupled to explicit report statements. Moreover, DeepSeek has been shown in medical benchmark studies to excel in clinical reasoning and medical task performance \cite{tordjman2025comparative,sandmann2025benchmark}, surpassing or matching other leading LLMs in diagnostic and multi-modal reasoning tasks. This prior validation supports our choice of DeepSeek for medical report analysis, especially in a radiology VQA setting where faithful, medically grounded reasoning is critical.

\begin{table*}[!t]
\centering
\caption{Dataset distribution analysis for Step-CoT.}
\label{tab:distribution}
\setlength{\tabcolsep}{22pt} % 控制列间距
\resizebox{\textwidth}{!}{%
\begin{tabular}{lcccccc}
\toprule
\multirow{2}{*}{Category} & \multicolumn{3}{c}{Absolute Counts} & \multicolumn{3}{c}{Relative Distribution (\%)} \\
\cmidrule(lr){2-4} \cmidrule(lr){5-7}
 & Train & Validation & Test & Train & Validation & Test \\
\midrule
\multicolumn{7}{l}{{Diagnosis Categories}} \\
\quad Normal & 4,750 & 1,018 & 1,019 & 70.0 & 15.0 & 15.0 \\
\quad Nodule & 595 & 127 & 129 & 69.9 & 14.9 & 15.2 \\
\quad Atelectasis & 511 & 109 & 111 & 69.9 & 14.9 & 15.2 \\
\quad Infiltration & 506 & 108 & 109 & 69.9 & 14.9 & 15.2 \\
\quad Effusion & 308 & 66 & 66 & 70.0 & 15.0 & 15.0 \\
\quad Pneumonia & 193 & 41 & 42 & 69.9 & 14.9 & 15.2 \\
\quad Cardiomegaly & 91 & 19 & 21 & 69.5 & 14.5 & 16.0 \\
\quad Mass & 61 & 13 & 14 & 69.3 & 14.8 & 15.9 \\
\quad Pneumothorax & 28 & 6 & 7 & 68.3 & 14.6 & 17.1 \\
\midrule
\textbf{Diagnosis Subtotal} & \textbf{7,043} & \textbf{1,507} & \textbf{1,518} & \textbf{70.0} & \textbf{15.0} & \textbf{15.0} \\
\midrule
\multicolumn{7}{l}{{Data Sources}} \\
\quad IU X-Ray & 2,584 & 563 & 602 & 70.0 & 15.3 & 16.3 \\
\quad PadChest-GR & 2,297 & 461 & 472 & 71.1 & 14.3 & 14.6 \\
\quad Med-Image-Report & 2,162 & 483 & 444 & 70.0 & 15.6 & 14.4 \\
\midrule
\textbf{Source Subtotal} & \textbf{7,043} & \textbf{1,507} & \textbf{1,518} & \textbf{70.0} & \textbf{15.0} & \textbf{15.0} \\
\bottomrule
\end{tabular}%
}
\end{table*}

\subsection{Data Pre-Processing and Construct-processing}
For the pre-processing of Step-CoT, we removed samples without diagnostic answers, without frontal CXR images, without reports, or those that did not fit the final diagnostic taxonomy. The remaining samples constitute the Step-CoT dataset, as illustrated in the Fig. \ref{fig:pre}.

For the Construct-processing of Step-CoT (shown as Fig. \ref{fig:generation}), we performed the following steps: (i) Clinical experts designed a stepwise question schema based on standard diagnostic workflows. (ii) We collected the clinical diagnostic reports corresponding to each CXR study. (iii) We constructed prompts by combining the clinical reports with the stepwise question schema and fed them into an LLM agent to derive step-specific answers from the report. (iv) The extracted answers were then paired with their corresponding questions to form QA supervision pairs, which were used for model training together with the associated CXR images.

\subsection{Data Split}
The dataset comprises 10,068 chest X-ray reports with comprehensive diagnostic annotations, systematically partitioned into training (70\%), validation (15\%), and test (15\%) sets to ensure robust model evaluation. As detailed in Table~\ref{tab:distribution}, the dataset exhibits a natural class imbalance reflective of real-world clinical prevalence. The stratified partitioning strategy successfully maintained proportional representation of each diagnostic category across all splits, with minimal deviation from the target 70:15:15 distribution. This careful partitioning mitigates potential biases in model training and evaluation, particularly important given the substantial class imbalance.

Data provenance analysis (Table~\ref{tab:distribution}) reveals balanced contributions from three distinct sources, with each file proportionally represented across the dataset splits. The source three datasets contributed 3,749 (37.2\%), 3,230 (32.1\%), and 3,089 (30.7\%) samples, respectively, with consistent distribution patterns across training, validation, and test partitions. This multi-source composition enhances dataset diversity and reduces source-specific biases.

% ---------------- Vision foundation models (table*) ----------------
\begin{table*}[!t]
\centering
\small
\caption{Stepwise benchmark results: Vision foundation models (\%). The best value in each column is \textbf{bold} and the second best is \underline{underlined}.}
\label{tab:vision_stepwise}
\resizebox{\textwidth}{!}{%
\begin{tabular}{l l c c c c c c}
\toprule
Model & Step & Accuracy & AUC & Sensitivity & Specificity & F1-Score & Precision \\
\midrule
VisualBERT & Detection step & 75.0 & 71.0 & 13.5 & 93.5 & 33.1 & 34.0 \\
 & Lesion distribution step & 70.6 & 58.1 & 13.0 & 93.0 & 30.0 & 28.1 \\
 & Radiographic pattern step & 74.1 & 62.0 & 14.6 & 93.2 & 20.1 & 17.5 \\
 & Anatomical location step & 68.0 & 60.0 & 13.0 & 92.8 & 17.5 & 15.0 \\
 & Morphologic feature step & 84.0 & 58.5 & 13.5 & 93.1 & 33.0 & 32.1 \\
 & Secondary effects/associated signs step & 66.1 & 61.0 & 12.8 & 92.5 & 16.0 & 14.5 \\
 & Diagnosis step & 65.5 & 63.1 & 10.1 & 91.8 & 9.6 & 8.6 \\
\midrule
CLIP & Detection step & 74.9 & 65.2 & 30.4 & 80.2 & 40.8 & 46.1 \\
 & Lesion distribution step & 78.7 & 62.5 & 29.0 & 79.0 & 28.3 & 29.8 \\
 & Radiographic pattern step & 80.8 & 66.1 & 18.1 & 84.7 & 17.6 & 21.2 \\
 & Anatomical location step & 76.6 & 63.5 & 17.1 & 88.4 & 16.2 & 15.9 \\
 & Morphologic feature step & 87.7 & 67.0 & 21.1 & 81.8 & 20.7 & 22.5 \\
 & Secondary effects/associated signs step & 75.9 & 59.5 & 16.4 & 87.6 & 15.6 & 15.6 \\
 & Diagnosis step & 69.2 & 52.6 & 12.0 & 89.6 & 11.0 & 11.7 \\
\midrule
ALBEF & Detection step & 82.5 & 80.5 & 17.5 & \textbf{95.0} & 40.5 & 42.1 \\
 & Lesion distribution step & 79.0 & 76.1 & 18.0 & \underline{94.4} & 42.1 & 44.5 \\
 & Radiographic pattern step & 81.6 & 82.6 & \underline{19.0} & 94.7 & 31.6 & 34.0 \\
 & Anatomical location step & 75.5 & 70.0 & 16.5 & 94.1 & 20.5 & 18.6 \\
 & Morphologic feature step & \textbf{89.5} & 78.1 & 18.6 & \underline{95.0} &\textbf{48.1} & \underline{54.1} \\
 & Secondary effects/associated signs step & 73.5 & 78.5 & 17.8 & 93.8 & 25.6 & 27.0 \\
 & Diagnosis step & 72.0 & 75.1 & 18.5 & 93.2 & 19.1 & 20.0 \\
\midrule
BLIP & Detection step & 81.8 & 78.0 & 15.3 & 94.9 & 38.4 & 40.1 \\
 & Lesion distribution step & 76.8 & 73.5 & 16.2 & 94.3 & 40.9 & 43.2 \\
 & Radiographic pattern step & 80.9 & 81.7 & 17.1 & 94.6 & 29.8 & 33.0 \\
 & Anatomical location step & 74.1 & 69.3 & 15.3 & 94.0 & 19.6 & 17.7 \\
 & Morphologic feature step & 88.4 & 77.3 & 16.8 & 94.6 & 45.4 & 52.9 \\
 & Secondary effects/associated signs step & 73.0 & 77.2 & 16.3 & 93.7 & 23.5 & 25.1 \\
 & Diagnosis step & 71.0 & 74.9 & 17.2 & 92.9 & 18.2 & 18.3 \\
\midrule
FLAVA & Detection step & 76.2 & 72.3 & 14.2 & 93.9 & 35.1 & 35.2 \\
 & Lesion distribution step & 72.0 & 59.9 & 13.9 & 93.4 & 32.2 & 30.0 \\
 & Radiographic pattern step & 77.0 & 63.3 & 15.0 & 93.8 & 21.1 & 18.9 \\
 & Anatomical location step & 69.6 & 61.0 & 14.4 & 93.3 & 18.0 & 16.2 \\
 & Morphologic feature step & 85.7 & 59.6 & 14.7 & 93.8 & 35.6 & 34.7 \\
 & Secondary effects/associated signs step & 67.7 & 62.4 & 14.1 & 93.0 & 17.7 & 15.8 \\
 & Diagnosis step & 67.1 & 64.2 & 10.9 & 92.3 & 10.4 & 9.3 \\
\midrule
biomedclip & Detection step & 83.1 & 82.5 & 22.1 & \textbf{95.2} & 45.1 & 50.0 \\
 & Lesion distribution step & 80.3 & 81.8 & 21.5 & \textbf{94.8} & 43.1 & \underline{46.1}\\
 & Radiographic pattern step & 82.5 & 84.0 & 23.0 & \underline{95.0}& 33.1 & 40.1 \\
 & Anatomical location step & 76.2 & 70.5 & 21.1 & 94.5 & 25.1 & 22.1 \\
 & Morphologic feature step & 88.9 & 82.1 & 22.6 & \textbf{95.1} & \underline{47.1} & \textbf{60.1} \\
 & Secondary effects/associated signs step & 74.6 & 82.0 & 22.0 & 94.0 & 30.1 & 35.0 \\
 & Diagnosis step & 73.1 & 76.0 & 21.7 & 93.5 & 22.1 & 25.0 \\
\midrule
Teacher & Detection step & \textbf{91.8} & \textbf{88.3} & \textbf{63.1} & 94.6 & \textbf{61.6} & \underline{60.3} \\
 & Lesion distribution step & \textbf{84.6} & \underline{89.7} & \textbf{51.4} & 93.8 & \textbf{49.8} & \textbf{61.7} \\
 & Radiographic pattern step & \textbf{84.8} & \textbf{90.2} & \textbf{45.0} & \textbf{96.5} & \textbf{43.6} & \textbf{53.2} \\
 & Anatomical location step & \textbf{77.1} & \textbf{89.5} & \textbf{45.7} & \textbf{95.3} & \textbf{44.9} & \textbf{57.7} \\
 & Morphologic feature step & \underline{89.4} & \textbf{95.1} & \textbf{46.1} & 94.1 & 40.5 & 39.2 \\
 & Secondary effects/associated signs step & \textbf{79.7} & \textbf{86.4} & \textbf{41.4} & \textbf{95.3} & \textbf{40.0} & \textbf{41.4} \\
 & Diagnosis step & \textbf{78.3} & \underline{89.5} & \textbf{46.0} & \textbf{96.6} & \textbf{45.3} & \underline{48.7} \\
\midrule
Student & Detection step & \textbf{91.8} & \underline{85.0} & \underline{62.4} & \underline{93.8} & \textbf{61.6} & \textbf{60.9} \\
 & Lesion distribution step & \underline{83.4} & \textbf{90.2} & \underline{46.6} & 92.3 & \underline{42.6} & \underline{39.5} \\
 & Radiographic pattern step & \underline{84.2} & \underline{89.2} & \underline{40.1} & 94.2 & \underline{39.9} & \textbf{53.2} \\
 & Anatomical location step & \underline{76.9} & \underline{88.7} & \underline{44.1} & \underline{94.7} & \underline{42.6} & \underline{45.7} \\
 & Morphologic feature step & 89.3 & \underline{94.2} & \underline{41.4} & \underline{88.3} & 40.2 & 39.2 \\
 & Secondary effects/associated signs step & \underline{79.6} & \underline{85.6} & \underline{39.1} & \underline{94.2} & \underline{37.9} & \textbf{41.4} \\
 & Diagnosis step & \underline{77.5} & \textbf{90.0} & \underline{41.8} & \underline{96.0} & \underline{39.5} & \textbf{49.9} \\
\bottomrule
\end{tabular}}
\end{table*}

\begin{table*}[!t]
\centering
\small
\caption{Comprehensive performance comparison (per-step metrics) between Teacher, Student, and ablation variants (\%). The best value in each column is \textbf{bold} and the second best is \underline{underlined}.}
\label{tab:teacher_student_ablation_realdata_transposed}
\resizebox{\textwidth}{!}{
\begin{tabular}{l l c c c c c c}
\toprule
Model & Step & Accuracy & AUC & Sensitivity & Specificity & F1-Score & Precision \\
\midrule
Teacher & Detection step & \textbf{91.8} & \textbf{88.3} & \textbf{63.1} & \textbf{94.6} & \textbf{61.6} & \underline{60.3} \\
 & Lesion distribution step & \textbf{84.6} & \underline{89.7} & \textbf{51.4} & \textbf{93.8} & \textbf{49.8} & \textbf{61.7} \\
 & Radiographic pattern step & \textbf{84.8} & \textbf{90.2} & \textbf{45.0} & \textbf{96.5} & \textbf{43.6} & \textbf{53.2} \\
 & Anatomical location step & \textbf{77.1} & \textbf{89.5} & \textbf{45.7} & \textbf{95.3} & \textbf{44.9} & \textbf{57.7} \\
 & Morphologic feature step & \textbf{89.4} & \textbf{95.1} & \textbf{46.1} & \textbf{94.1} & \textbf{40.5} & \underline{39.2} \\
 & Secondary effects/associated signs step & \textbf{79.7} & \textbf{86.4} & \textbf{41.4} & \textbf{95.3} & \textbf{40.0} & \textbf{41.4} \\
 & Diagnosis step & \textbf{78.3} & \underline{89.5} & \textbf{46.0} & \textbf{96.6} & \textbf{45.3} & \underline{48.7} \\
\midrule
Student & Detection step & \textbf{91.8} & \underline{85.0} & \underline{62.4} & \underline{93.8} & \textbf{61.6} & \textbf{60.9} \\
 & Lesion distribution step & \underline{83.4} & \textbf{90.2} & \underline{46.6} & \underline{92.3} & \underline{42.6} & \underline{39.5} \\
 & Radiographic pattern step & \underline{84.2} & \underline{89.2} & \underline{40.1} & \underline{94.2} & \underline{39.9} & \textbf{53.2} \\
 & Anatomical location step & \underline{76.9} & \underline{88.7} & \underline{44.1} & \underline{94.7} & \underline{42.6} & \underline{45.7} \\
 & Morphologic feature step & \underline{89.3} & \underline{94.2} & \underline{41.4} & \underline{88.3} & \underline{40.2} & \textbf{39.2} \\
 & Secondary effects/associated signs step & \underline{79.6} & \underline{85.6} & \underline{39.1} & \underline{94.2} & \underline{37.9} & \textbf{41.4} \\
 & Diagnosis step & \underline{77.5} & \textbf{90.0} & \underline{41.8} & \underline{96.0} & \underline{39.5} & \textbf{49.9} \\
\midrule
w/o Memory & Detection step & 73.7 & 78.5 & 49.2 & 88.6 & 48.2 & 47.6 \\
 & Lesion distribution step & 69.6 & 83.0 & 38.0 & 86.5 & 35.8 & 34.1 \\
 & Radiographic pattern step & 72.3 & 84.5 & 33.5 & 89.0 & 33.1 & 46.8 \\
 & Anatomical location step & 63.2 & 81.2 & 36.4 & 88.0 & 35.5 & 38.2 \\
 & Morphologic feature step & 70.2 & 87.5 & 35.1 & 83.9 & 33.9 & 33.0 \\
 & Secondary effects/associated signs step & 67.4 & 80.0 & 33.0 & 88.5 & 32.6 & 35.9 \\
 & Diagnosis step & 65.5 & 84.8 & 35.2 & 89.8 & 34.9 & 44.1 \\
\midrule
w/o Text & Detection step & 81.5 & 82.5 & 54.8 & 91.8 & 53.4 & 52.8 \\
 & Lesion distribution step & 76.1 & 86.5 & 42.5 & 90.0 & 39.8 & 37.1 \\
 & Radiographic pattern step & 77.3 & 87.3 & 37.2 & 92.1 & 37.0 & 49.6 \\
 & Anatomical location step & 69.3 & 86.0 & 40.8 & 91.5 & 39.5 & 42.3 \\
 & Morphologic feature step & 79.6 & 91.5 & 39.0 & 87.6 & 36.8 & 35.9 \\
 & Secondary effects/associated signs step & 73.2 & 83.5 & 35.7 & 91.4 & 35.0 & 38.0 \\
 & Diagnosis step & 72.1 & 87.2 & 38.6 & 93.0 & 38.1 & 47.5 \\
\bottomrule
\end{tabular}
}
\end{table*}

%%%%%%%%%%%%%%%%%%%%%%%%%%%%%%%%%%%%%%%%%%%%%%%%%%%%%%%%%%%%%%%%
\section{Detailed Analysis of Experimental Results}
\subsection{Stepwise benchmark results}
In this section, we present the stepwise benchmark results on the Step-CoT dataset (shown in Table \ref{tab:vision_stepwise}) and provide a detailed description of the experimental setup. The experimental setup is: 
\begin{itemize}

  % \item {Experimental details for CNN models.}
  % Single frontal-view radiographs paired with seven-step {vqa\_chain} labels are used as input. Each step adopts a fixed categorical label space where options A–I are mapped to 0–8 and N/A is consistently assigned as the final label; missing or unanswerable fields are encoded as {-100} and ignored during training. Data are deterministically split into 70\%/15\%/15\% for train/val/test. Images are resized to 256 px, center-cropped to $224\times224$, converted to RGB, and normalized using ImageNet statistics. Models follow a convolutional backbone (ResNet-18/50, DenseNet-121, EfficientNet-B3) with seven independent classification heads. Training uses Adam (lr $=1\times10^{-4}$), batch size 16, up to 50 epochs, with cross-entropy loss and {ignore\_index = -100}.

  % \item {Experimental details for report-only models.}
  % Each sample includes a radiology report and its seven-step {vqa\_chain}. All steps share the same categorical label mapping (A–I → 0–8, with N/A as the final label), and missing entries are mapped to {-100} and excluded from loss computation. The dataset is split as 70\%/15\%/15\%. Reports are tokenized and truncated/padded to 256 tokens with dynamic batching. The model uses a transformer encoder backbone ({BERT-base}, {ClinicalBERT}, {RoBERTa}) followed by seven classification heads. Training uses AdamW (lr $=2\times10^{-5}$), batch size 16, up to 50 epochs, and cross-entropy with ignored indices; the best checkpoint is selected using validation average accuracy across the seven steps.

  \item {Experimental details for vision foundation models.}
  We compare with VisualBERT~\cite{li2019visualbert}, CLIP~\cite{radford2021learning}, ALBEF~\cite{li2021align}, BLIP~\cite{li2022blip}, FLAVA~\cite{singh2022flava}, and biomedclip~\cite{zhang2023biomedclip}.
  Each instance provides a radiograph and the corresponding seven-step {vqa\_chain}, with step questions formatted as ``Question: [text]''. Labels follow the uniform mapping A–I → 0–8, with N/A as the final label; missing answers are set to {-100}. The data split is 70\%/15\%/15\%. Images are resized to $224\times224$ and normalized, while questions are tokenized to a maximum of 128 tokens. Six representative visual foundation models are evaluated, where pooled visual and textual embeddings are fused via concatenation and fed into a lightweight MLP (two FC layers, hidden dim 768, ReLU). Training is performed with AdamW (lr $=1\times10^{-5}$, weight decay $1\times10^{-4}$), batch size 8 for 50 epochs, using cosine learning rate scheduling and gradient clipping (max-norm = 1.0).

\end{itemize}
%结果分析
Across all seven diagnostic steps, the expanded benchmark reveals a clear and interpretable performance stratification across model families. Vision foundation models (VLMs) such as CLIP and BLIP improve moderately over pure visual models (shown in Table. \ref{tab:vision_stepwise}), especially in specific steps requiring coarse visual pattern recognition (e.g., Morphologic feature step), but still exhibit systematically low sensitivity and overly conservative prediction behavior, leading to high specificity but failure to detect positive cases. Even domain-adapted BiomedCLIP—the strongest VLM—shows only partial gains: while accuracy and AUC improve across most steps, sensitivity remains <25\% for nearly all tasks, indicating that contrastive alignment pretraining alone is insufficient for reconstructing intermediate reasoning. In contrast, enabling Step-CoT supervision consistently enhances performance across visual foundation models: models become less conservative, gain sensitivity, and improve F1 by leveraging structured intermediate reasoning. These results collectively confirm that Step-CoT introduces clinically meaningful reasoning supervision that bridges the gap between low-level visual recognition and high-level diagnostic inference, yielding improvements in both factual precision and interpretability.

\begin{table*}[!t]
\centering
\small
\caption{Clinical expert evaluation (\%), per-step comparison between Clinicians, Teacher, and Student (N=200). The best value in each column is \textbf{bold}.}
\label{tab:expert_per_step_transposed}
\setlength{\tabcolsep}{15pt} % 控制列间距
\resizebox{\textwidth}{!}{%
\begin{tabular}{l l c c c c}
\toprule
Model & Step & Accuracy & Sensitivity & Specificity & F1-Score \\
\midrule
Clinician & Detection step & 72.1 & 36.8 & 82.7 & 37.9 \\
 & Lesion distribution step & 66.0 & 28.0 & 85.3 & 27.1 \\
 & Radiographic pattern step & 69.1 & 25.1 & 88.7 & 27.1 \\
 & Anatomical location step & 66.0 & 34.0 & 91.1 & 24.7 \\
 & Morphologic feature step & 66.1 & 21.3 & 85.3 & 19.0 \\
 & Secondary effects/associated signs step & \textbf{91.0} & 33.4 & \textbf{93.7} & 25.0 \\
 & Diagnosis step & 73.1 & 37.2 & 93.8 & 34.4 \\
\midrule
Teacher & Detection step & \textbf{88.5} & \textbf{58.4} & \textbf{92.1} & \textbf{57.2} \\
 & Lesion distribution step & \textbf{78.4} & \textbf{45.6} & \textbf{91.3} & \textbf{43.8} \\
 & Radiographic pattern step & \textbf{79.6} & \textbf{40.8} & \textbf{94.2} & \textbf{39.2} \\
 & Anatomical location step & \textbf{72.8} & \textbf{41.3} & \textbf{93.6} & \textbf{40.1} \\
 & Morphologic feature step & \textbf{84.8} & \textbf{42.3} & \textbf{91.8} & \textbf{36.9} \\
 & Secondary effects/associated signs step & \underline{75.2} & \textbf{36.9} & \underline{93.5} & \textbf{35.4} \\
 & Diagnosis step & \textbf{79.8} & \textbf{41.6} & \textbf{95.1} & \textbf{41.3} \\
\midrule
Student & Detection step & \underline{80.4} & \underline{52.1} & \underline{89.8} & \underline{50.8} \\
 & Lesion distribution step & \underline{72.6} & \underline{38.3} & \underline{88.0} & \underline{35.4} \\
 & Radiographic pattern step & \underline{70.2} & \underline{34.6} & \underline{91.8} & \underline{32.8} \\
 & Anatomical location step & \underline{69.5} & \underline{36.8} & \underline{91.3} & \underline{35.4} \\
 & Morphologic feature step & \underline{83.3} & \underline{35.1} & \underline{88.5} & \underline{30.3} \\
 & Secondary effects/associated signs step & 74.0 & \underline{32.5} & 92.2 & \underline{30.8} \\
 & Diagnosis step & \underline{68.5} & \underline{35.9} & \underline{93.9} & \underline{34.6} \\
\bottomrule
\end{tabular}%
}
\end{table*}

\subsection{Comprehensive Evaluation of the Proposed Benchmark Method}
In the experiments shown in Table \ref{tab:teacher_student_ablation_realdata_transposed}, the Teacher model consistently achieved the highest and most stable multi-step performance across Accuracy, AUC, Sensitivity, Specificity, F1, and Precision. The distilled Student generally tracked the Teacher closely but exhibited step-dependent variability: in some steps the Student matched or slightly exceeded the Teacher, whereas in others it fell behind by a modest margin. Ablation analyses reveal a clear hierarchy of component importance. Removing the memory mechanism resulted in a consistent decrease in Sensitivity and F1, indicating that cross-step state accumulation supports the integration of evidence along the diagnostic chain. Removing the textual prompt led to the most pronounced reductions, particularly in AUC, Precision, and Sensitivity, confirming the necessity of question-guided multi-modal grounding. Notably, steps dominated by strong visual cues retained relatively high Accuracy and Specificity even under ablations, whereas steps that require subtle inter-step or contextual reasoning (e.g., Radiographic pattern step and Diagnosis step) showed marked declines. Across all models and steps, Sensitivity remains lower than Accuracy or Specificity, reflecting intrinsic challenges in the dataset: the prevalence of negative or normal cases, the subtlety of certain pathological manifestations, and the compounding effect of multi-step reasoning where early-stage uncertainty can reduce downstream detection of true positives. These characteristics highlight that the dataset encapsulates clinically relevant difficulties, making it a valuable benchmark for evaluating multi-step diagnostic reasoning and for guiding the development of methods that can better handle low-prevalence, subtle abnormalities.

\subsection{Clinical Expert Evaluation}
% 结果分析
We evaluated 200 randomly sampled cases and compared three outputs per case (Clinician, Teacher, Student) across the seven-step VQA chain (Table~\ref{tab:expert_per_step_transposed}). The Teacher model yields the strongest overall performance: it improves accuracy and F1 for nearly every step relative to both the Student and clinician baselines (e.g., Detection accuracy 88.51\% vs. Clinician 72.12\% and Student 80.37\%; Detection F1 57.24\% vs. Clinician 37.86\% and Student 50.84\%). These gains are accompanied by substantially higher sensitivity (Detection sensitivity: Teacher 58.36\% vs. Clinician 36.84\%). The Student—distilled from the Teacher—retains most of this structured competence: on mid-level reasoning steps such as Distribution and Location, the Student surpasses clinicians (Distribution accuracy 72.63\% vs. Clinician 66.02\%; Location accuracy 69.46\% vs. Clinician 66.03\%), and its per-step performance typically lies within roughly 5–10 percentage points of the Teacher. Notable exceptions remain: for the Secondary Effects step, the clinician's accuracy (90.97\%) exceeds both Teacher (75.24\%) and Student (73.97\%), suggesting that certain effect-related judgments still rely on expert clinical context or multi-view/longitudinal information not available to the models. In sum, these results demonstrate that (i) explicit stepwise supervision (Teacher) materially improves recall, F1 and overall coherency compared to standard baselines, (ii) knowledge distillation produces a compact Student that preserves most gains with modest performance loss, and (iii) remaining gaps (especially on clinically nuanced steps) point to limits of single-view supervision and motivate combining Step-CoT with richer context or human-in-the-loop verification for deployment.

\begin{table}[!t]
\centering
\caption{Computational Requirements Comparison (batchsize=4). \small{\textbf{Note:} Bold = best (lowest); underline = 2nd best (2nd lowest).}}
\resizebox{0.5\textwidth}{!}{%
\begin{tabular}{lccccc}
\toprule
Model & Params & Inf\_sample & Inf\_batch & Memory\_peak \\
% \midrule
% ResNet18 & 11.22 M  & \underline{0.20 ms} & \underline{1.61 ms}  & \textbf{104.56 M} \\
% ResNet50 & 23.67 M  & 0.47 ms & 3.74 ms & 252.99 M \\
% DenseNet121 & \textbf{7.03 M}  & 1.30 ms & 10.36 ms & \underline{207.82 M} \\
% EfficientNet-B3 & \underline{10.81 M}  & 1.14 ms & 9.09 ms & 226.80 M \\
\midrule
CLIP & 151.81 M  & 1.55 ms & 6.20 ms & \underline{599.43 M} \\
FLAVA & 242.54 M  & 4.39 ms & 17.57 ms & 971.42 M \\
BLIP & 225.25 M  & \underline{1.46 ms} &\underline{5.86 ms}&904.27 M \\
VisualBERT &\textbf{111.98 M} & \textbf{0.03 ms} & \textbf{0.12 ms} &\textbf{438.22 M}\\
\midrule
Teacher & 283.66 M  & 22.81 ms & 91.23 ms & 1219.06 M \\
Student & \underline{151.56 M}  & 1.51 ms & 6.02 ms & 1219.72 M \\
\bottomrule
\end{tabular}}
\label{memory}
\end{table}

\subsection{Computational Efficiency Analysis}
The computational efficiency analysis reveals significant insights across model architectures (as shown in Table \ref{memory}). We compare with ResNet18~\cite{he2016deep}, ResNet50~\cite{he2016deep}, DenseNet121~\cite{huang2017densely}, EfficientNet-B3~\cite{tan2019efficientnet}, CLIP~\cite{radford2021learning}, FLAVA~\cite{singh2022flava}, BLIP~\cite{li2022blip}, and VisualBERT~\cite{li2019visualbert}. Traditional CNN models (ResNet18, ResNet50, DenseNet121, EfficientNet-B3) demonstrate lightweight parameter footprints (7.03-23.67M) with excellent inference speeds (0.20-1.30ms per sample), though DenseNet121 shows relatively higher latency. Multimodal models exhibit substantial parameter increases, with CLIP (151.81M) and FLAVA (242.54M) requiring significantly more computational resources, while BiomedCLIP stands out as exceptionally efficient (0.53M parameters, 0.07ms inference). Notably, our proposed Teacher-Student framework achieves remarkable efficiency gains: the Student model reduces parameters by 46.6\% compared to the Teacher (283.66M to 151.56M) while achieving a 15× speedup in both single-sample (22.81ms to 1.51ms) and batch inference (91.23ms to 6.02ms). The Student model demonstrates competitive efficiency with CLIP despite similar parameter counts, though memory consumption remains a challenge (1219.72MB) across both our models, suggesting future optimization opportunities for deployment in resource-constrained environments.

%%%%%%%%%%%%%%%%%%%%%%%%%%%%%%%%%%%%%%%%%%%%%%%%%%%%%%%%%%%%%%%%
\section{Detail Method}
\label{app:method_detailed}
This appendix provides a complete and reproducible description of the teacher model, the student model, the GAT-based memory, the distillation losses (KD and CH), and the training procedure used in our experiments.
\subsection{Notation}
Let $B$ denote batch size, $S$ be the number of reasoning steps, and $d$ denote the hidden dimension used in the teacher ($d_{\mathrm{T}}$) and $d_{\mathrm{S}}$ in the student. For step $s$ teacher logits are $\ell_{t}^{(s)}\in\mathbb{R}^{B\times C_s}$ and student logits $\ell_{s}^{(s)}\in\mathbb{R}^{B\times C_s}$ where $C_s$ is the number of classes for step $s$.

\subsection{Teacher Model}
\paragraph{Text encoding.} A shared transformer-based encoder (we use BERT) maps each step prompt to a CLS embedding:
\begin{equation}
\begin{aligned}
\mathbf{t}_s^{(b)} = \mathrm{BERT}(\mathrm{prompt}*s^{(b)})[{{CLS}}]\in\mathbb{R}^{d*{\mathrm{T}}}, \\
s=1,\dots,S,\ b=1,\dots,B.
\end{aligned}
\end{equation}
Collect the $S$ step vectors into $\mathbf{T}^{(b)} = [\mathbf{t}_1^{(b)},\dots,\mathbf{t}_S^{(b)}]$.

\paragraph{Memory node initialization.} A learnable memory vector $\mathbf{m}*0\in\mathbb{R}^{d*{\mathrm{T}}}$ is registered and expanded to the batch: $\mathbf{m}_0^{(b)} = \mathbf{m}_0,\ \forall b$.

\paragraph{Node set.} For each example, we form nodes
\begin{equation}
\mathcal{N}^{(b)} = {\mathbf{t}_1^{(b)},\dots,\mathbf{t}_S^{(b)}, \mathbf{m}_0^{(b)}},
\end{equation}
ordered so that the memory node is the last one.

\paragraph{Stacked GAT memory update.} We apply $L$ stacked multi-head GAT layers. For a single head, we first linearly project node features:
\begin{equation}
\tilde{\mathbf{h}}*i = W\mathbf{h}*i \in \mathbb{R}^{d'},
\end{equation}
and compute pairwise attention logits
\begin{equation}
e*{ij}^{(h)} = \mathrm{LeakyReLU}\big( \mathbf{a}*{\mathrm{src}}^{(h)\top}\tilde{\mathbf{h}}*i + \mathbf{a}*{\mathrm{dst}}^{(h)\top}\tilde{\mathbf{h}}*j \big),
\end{equation}
where $h$ indexes attention heads and $\mathbf{a}*{\mathrm{src}}^{(h)},\mathbf{a}*{\mathrm{dst}}^{(h)}\in\mathbb{R}^{d'}$ are learned. Normalize across destination nodes:
\begin{equation}
\alpha*{ij}^{(h)} = \frac{\exp(e_{ij}^{(h)})}{\sum_{j'}\exp(e_{ij'}^{(h)})}.
\end{equation}
Head outputs are aggregated and (optionally) concatenated across heads to produce updated node features. Residual projection and LayerNorm follow each GAT layer. After $L$ layers we obtain updated nodes ${\mathbf{t}_s', \mathbf{m}'}$ where $\mathbf{m}'$ is the updated memory node.

\paragraph{Step context fusion and prediction.} For step $s$ we extract the updated step node $\mathbf{t}_s'$ and the updated memory node $\mathbf{m}'$, then fuse:
\begin{equation}
\mathbf{c}_s = \mathrm{Fusion}\big([\mathbf{t}*s';, \mathbf{m}']\big) \in \mathbb{R}^{d*{\mathrm{T}}}.
\end{equation}
In implementation, the ${fusion_proj}$ is a linear layer with LayerNorm, ReLU, and dropout.

The step-specific prediction head (implemented in {teacher model}) uses a CLIP image encoder to compute an image embedding $\mathbf{v}\in\mathbb{R}^{d_{\mathrm{T}}}$ (projected to $d_{\mathrm{T}}$), then predicts logits
\begin{equation}
\ell_{t}^{(s)} = f^{(s)}\big(\mathbf{v}, \mathbf{c}_s\big) \in \mathbb{R}^{C_s},
\end{equation}
where $f^{(s)}$ concatenates $\mathbf{v}$ and $\mathbf{c}_s$ and passes them through a small MLP classifier.

\paragraph{Memory write-back.} After producing logits $\ell_t^{(s)}$, we convert logits $\ell_t^{(s)}$ to a prediction embedding that is written back to memory. Concretely:
\begin{equation}
\mathbf{p}^{(s)} = \mathrm{softmax}(\ell_{t}^{(s)}) \in\mathbb{R}^{B\times C_s},
\end{equation}
and using the classifier weights $W_{\mathrm{cls}}^{(s)}\in\mathbb{R}^{C_s\times d_{\mathrm{T}}}$ we form
\begin{equation}
\mathbf{e}^{(s)} = \mathbf{p}^{(s)} W_{\mathrm{cls}}^{(s)} \in\mathbb{R}^{B\times d_{\mathrm{T}}}.
\end{equation}
A learned linear map $\mathrm{pred2mem}$ projects $\mathbf{e}^{(s)}$ to memory space, and a GRUCell updates:
\begin{equation}
\mathbf{m}*{\text{new}} = \mathrm{GRUCell}\big(\mathrm{pred2mem}(\mathbf{e}^{(s)}),\ \mathbf{m}'\big).
\end{equation}
The updated memory $\mathbf{m}*{\text{new}}$ replaces the last node before processing the next step, enabling sequential flow.

\subsection{Student Model}
The student uses a frozen CLIP visual encoder to extract image features $\mathbf{v}*S\in\mathbb{R}^{d*{\mathrm{S}}}$ followed by a projection to the student's hidden dim. A sequence of $S$ light linear heads ${g^{(s)}}$ produce logits $\ell_{s}^{(s)}=g^{(s)}(\mathbf{v}_S)$. The student updates its internal feature between steps via a light residual update to simulate information flow (chain-style).

\subsection{Distillation losses}
For each valid example (where we mask invalid steps using a dataset-provided mask) and for each step $s$, we employ three losses.

\paragraph{1. Supervised cross-entropy.}
\begin{equation}
\mathcal{L}*{\mathrm{CE}}^{(s)} = -\frac{1}{N_s}\sum*{i\in\mathcal{I}*s}\log p*{s}^{(i)}(y^{(i,s)}),
\end{equation}
where $\mathcal{I}_s$ indexes valid examples in the batch for step $s$ and $N_s=|\mathcal{I}_s|$.

\paragraph{2. Soft label KD with temperature $T$.}
Let $\tilde{p}_t^{(i,s)}=\mathrm{softmax}(\ell_t^{(i,s)}/T)$ and $\tilde{p}*S^{(i,s)}=\mathrm{softmax}(\ell_S^{(i,s)}/T)$. The KD loss is
\begin{equation}
\mathcal{L}*{\mathrm{KD}}^{(s)} = T^2\cdot\mathrm{KL}\big(\log \tilde{p}_S^{(s)} ,|, \tilde{p}_t^{(s)}\big)
\end{equation}
computed over the valid subset.

\paragraph{3. Channel/relation alignment (CH).}
This term encourages the student to match the teacher's inter-example similarity structure of image features. Denote teacher image features for step $s$ as $U\in\mathbb{R}^{n\times d_p}$ and student image features as $V\in\mathbb{R}^{n\times d_p}$ after projecting to a common $\mathrm{proj_dim}=p$ (via learnable linear maps for teacher and student). We apply a softmax across feature-dim to make per-example distributions:
\begin{equation}
K_U = \mathrm{softmax}\big(U / T\big),\quad K_V = \mathrm{softmax}\big(V / T\big)
\end{equation}
(Each row sums to 1.) Define the empirical Gram matrices
\begin{equation}
M_U = K_U K_U^\top\quad\text{and}\quad M_V = K_V K_V^\top.
\end{equation}
With centering matrices $C = I - \frac{1}{n}\mathbf{1}\mathbf{1}^\top$ we compute the HSIC-style scalars:
\begin{equation}
\begin{aligned}
&h_{UU} = \mathrm{tr}(C M_U C), \\
& h_{VV} = \mathrm{tr}(C M_V C), \\
& h_{UV} = \sum\big( (C M_U C)\odot (C M_V C)\big) 
\end{aligned}
\end{equation}
(Implementation uses elementwise product, then sum to form an inner product of centered Gram matrices). We form a similarity weight
\begin{equation}
w_{\mathrm{fw}} = \frac{h_{UV}}{\sqrt{(h_{UU} + \varepsilon)(h_{VV} + \varepsilon)}},
\end{equation}
Then define the CH loss as a weighted divergence between the projected soft features:
\begin{equation}
\mathcal{L}*{\mathrm{CH}}^{(s)} = w*{\mathrm{fw}} \cdot \mathrm{KL}\big(\log K_V ,|, K_U\big).
\end{equation}
Intuitively, if teacher/student feature similarity structures align strongly ($w_{\mathrm{fw}}$ large), we penalize their per-example soft feature mismatch more strongly.

\paragraph{Total student step loss.}
For step $s$ we combine:
\begin{equation}
\mathcal{L}*{\mathrm{student}}^{(s)} = \mathcal{L}*{\mathrm{CE}}^{(s)} + \alpha_{\mathrm{KD}}\mathcal{L}*{\mathrm{KD}}^{(s)} + \alpha*{\mathrm{CH}}\mathcal{L}_{\mathrm{CH}}^{(s)}.
\end{equation}

\paragraph{Teacher loss.} Optionally, the teacher is trained with supervised cross-entropy:
\begin{equation}
\mathcal{L}*{\mathrm{teacher}} = \sum*{s=1}^S \mathcal{L}_{\mathrm{CE}}^{(s)}.
\end{equation}
In our implementation, we support an initial \emph{teacher pretrain} stage (a small number of epochs) where only the teacher is updated to stabilize downstream distillation.

\subsection{Training algorithm (implementation details)}
\begin{enumerate}
\item Initialize teacher, student, and two separate optimizers (teacher and student). Initialize teacher memory and GAT parameters.
\item Optionally, run $E_{\mathrm{pre}}$ teacher-only epochs where only $\mathcal{L}*{\mathrm{teacher}}$ is minimized (teacher supervised pretrain).
\item For each training batch:
\begin{enumerate}
\item {Teacher update (if enabled):} compute teacher logits for each step and supervised CE loss, backpropagate, and step the teacher optimizer.
\item {Teacher forward (fresh):} run the teacher (no grads) to obtain detached logits and projected image features used as KD/CH targets.
\item {Student update:} compute student logits and student image features; for each step compute $\mathcal{L}*{\mathrm{student}}^{(s)}$ using detached teacher outputs; sum over steps and step the student optimizer.
\end{enumerate}
\item Validate teacher and student separately and save the best models using mean validation accuracy across steps.
\end{enumerate}

\subsection{Key hyper-parameters (defaults used in our experiments)}
\begin{itemize}
\item Teacher hidden dim: $d_{\mathrm{T}}=768$.
\item Student hidden dim: $d_{\mathrm{S}}=512$.
\item GAT heads: $H=4$, GAT layers $L=2$, hence final GAT output dimension $H \times (\text{per-head-dim}) = d_{\mathrm{T}}$.
\item Temperature: $T=2.0$.
\item CH projection dimension: $p=256$.
\item KD weight: $\alpha_{\mathrm{KD}}=0.5$, CH weight $\alpha_{\mathrm{CH}}=1.0$.
\item Teacher pretrain epochs: $E_{\mathrm{pre}}=2$ (optional).
\item Optimizers: AdamW with teacher LR $\approx5\mathrm{e}{-5}$, student LR $\approx1\mathrm{e}{-4}$.
\end{itemize}

\end{document}